\PassOptionsToPackage{table}{xcolor}   % must be the very first thing
\pdfoutput=1
\documentclass[11pt]{article}

\usepackage[preprint]{acl}

\usepackage{times}
\usepackage{latexsym}

\usepackage[T1]{fontenc}
\usepackage[utf8]{inputenc}

\usepackage{microtype}

\usepackage{inconsolata}

\usepackage{graphicx}
\usepackage{amsmath}
\usepackage{mathrsfs}
\usepackage{booktabs}    
\usepackage{amssymb}
\usepackage{amsthm}
\usepackage{subcaption}
\usepackage{mathtools}
\usepackage{multirow}
\usepackage{float}
\usepackage{hyperref}
\usepackage{xcolor}
\usepackage{amsmath,amsfonts,bm}

\def\eqref#1{equation~\ref{#1}}
\def\1{\bm{1}}

\def\vg{{\bm{g}}}
\def\vh{{\bm{h}}}

\DeclareMathAlphabet{\mathsfit}{\encodingdefault}{\sfdefault}{m}{sl}
\SetMathAlphabet{\mathsfit}{bold}{\encodingdefault}{\sfdefault}{bx}{n}

\newcommand{\sigmoid}{\sigma}

\usepackage{hyperref}
\usepackage{url}
\usepackage{multirow}
\usepackage[most]{tcolorbox}
\usepackage{graphicx}
\usepackage{wrapfig}
\usepackage[table]{xcolor}
\usepackage{longtable}

\usepackage[utf8]{inputenc} % allow utf-8 input
\usepackage[T1]{fontenc}    % use 8-bit T1 fonts
\usepackage{hyperref}       % hyperlinks
\usepackage{url}            % simple URL typesetting
\usepackage{booktabs}       % professional-quality tables
\usepackage{amsfonts}       % blackboard math symbols
\usepackage{nicefrac}       % compact symbols for 1/2, etc.
\usepackage{microtype}      % microtypography
\usepackage{xcolor}         % colors
\usepackage{standalone}
\usepackage{latexsym}
\usepackage{amsmath}
\usepackage{amssymb}
\usepackage{amsthm}
\usepackage{natbib}   
\usepackage{graphicx}
\usepackage{subcaption}
\usepackage{array}
\usepackage{tabu}
\usepackage{makecell}
\usepackage{paralist}
\usepackage{cases}
\usepackage{diagbox}
\usepackage{enumitem}
\usepackage{soul}
\usepackage{multirow}
\usepackage{verbatim}
\usepackage{tabulary}
\usepackage{booktabs}
\usepackage{tabularx}
\usepackage[mathscr]{euscript}
\usepackage{mathtools}
\usepackage{algorithm}
\usepackage{algpseudocode}
\usepackage{stmaryrd}
\usepackage{tikz-dependency}
\usetikzlibrary{automata,decorations.markings,arrows,positioning,matrix,calc,patterns,angles,quotes,calc}
\usepackage{adjustbox}
\usepackage{tabularx}
\usepackage{xspace}
\usepackage{tabulary}
\usepackage{afterpage}
\usepackage{bm}
\usepackage{color}
\usepackage{graphicx}
\usepackage{slashbox}
\usepackage[toc,page]{appendix}
\usepackage{makecell}
\usepackage{boldline}
\usepackage[shortcuts]{extdash}  % for non-breaking hyphens. Stack Overflow says it's important for this to be the last package loaded.

\usepackage{blindtext}
\usepackage{graphicx}
\usepackage{capt-of}
\usepackage{booktabs}
\usepackage{varwidth}
\usepackage{pifont}% http://ctan.org/pkg/pifont
\usepackage{wrapfig}

\usepackage{listings}

\usepackage{pythonhighlight}

\definecolor{orange}{rgb}{1,0.5,0}
\definecolor{mdgreen}{rgb}{0.05,0.6,0.05}
\definecolor{mdblue}{rgb}{0,0,0.7}
\definecolor{dkblue}{rgb}{0,0,0.5}
\definecolor{dkgray}{rgb}{0.3,0.3,0.3}
\definecolor{slate}{rgb}{0.25,0.25,0.4}
\definecolor{gray}{rgb}{0.5,0.5,0.5}
\definecolor{ltgray}{rgb}{0.7,0.7,0.7}
\definecolor{purple}{rgb}{0.7,0,1.0}
\definecolor{lavender}{rgb}{0.65,0.55,1.0}

\definecolor{mypurple}{RGB}{111,61,121}
\definecolor{myblue}{RGB}{46,88,180}
\definecolor{myred}{RGB}{181,68,106}
\definecolor{myyellow}{RGB}{204,143,55}

\newcommand{\term}[1]{\textbf{#1}} % term being defined

\newcommand{\relu}{\operatorname{ReLU}}
\newcommand{\classifier}{$f_\theta$\xspace}
\newcommand{\intervention}{$\mathbf{g}_\phi$\xspace}

\DeclareSymbolFont{extraup}{U}{zavm}{m}{n}
\DeclareMathSymbol{\vardiamond}{\mathalpha}{extraup}{87}

\newcolumntype{L}[1]{>{\raggedright\let\newline\\\arraybackslash\hspace{0pt}}m{#1}}
\newcolumntype{C}[1]{>{\centering\let\newline\\\arraybackslash\hspace{0pt}}m{#1}}
\newcolumntype{R}[1]{>{\raggedleft\let\newline\\\arraybackslash\hspace{0pt}}m{#1}}

\theoremstyle{definition}

\theoremstyle{remark}

\algrenewcommand{\algorithmiccomment}[1]{\leavevmode$\triangleright$ #1}

\setul{1pt}{.4pt}

\newcommand*{\name}{\textsc{FactCheckmate}\xspace}

\DeclareFixedFont{\ttb}{T1}{txtt}{bx}{n}{12} % for bold
\DeclareFixedFont{\ttm}{T1}{txtt}{m}{n}{12}  % for normal

\usepackage{subcaption}
\title{
\raisebox{-0.1\height}{\includegraphics[width=0.6cm,height=0.6cm,keepaspectratio]{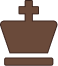}}
\name: Preemptively Detecting and Mitigating Hallucinations in LMs
}

\author{Deema Alnuhait$^{1}$\thanks{Equal contribution.} \hspace{2pt}, 
Neeraja Kirtane$^{1}$\footnotemark[1] \hspace{2pt}, 
Muhammad Khalifa$^{2}$, \textbf{Hao Peng}$^{1}$ \\ 
$^1$University of Illinois Urbana-Champaign, 
$^2$University of Michigan \\ 
{\texttt{\{deemaa2,kirtane3,haopeng\}@illinois.edu}}\\
{\texttt{khalifam@umich.edu}} 
}

\usepackage{colortbl}   % defines \cellcolor and \columncolor

\begin{document}
\maketitle
\begin{abstract}
% The abstract paragraph should be indented 1/2~inch (3~picas) on both left and
% right-hand margins. Use 10~point type, with a vertical spacing of 11~points.
% The word \textsc{Abstract} must be centered, in small caps, and in point size 12. Two
% line spaces precede the abstract. The abstract must be limited to one
% paragraph.
% \hao{taking a stab on the opening}
% \hao{Language models hallucinate.
% We inquire:
% can their nonfactual generations be detected and mitigated \emph{before} they are even generated?
% }
% \mk{I would rephrase a bit: can we preemtively mitigate such hallucination before it occurs?}
%With \neeraja{our framework}
% \textbf{Notes: Improve the narrative of the preemptive part.}
Language models (LMs) hallucinate.
We inquire:
Can we detect and mitigate hallucinations \emph{before} they happen?
This work answers this research question in the positive, by showing that the internal representations of LMs provide rich signals that can be used for this purpose.
We introduce \name, which \textit{preemptively} detects hallucinations by learning a classifier that predicts whether the LM \emph{will} hallucinate, based on the model's hidden states produced over the inputs, \emph{before} decoding begins.
If a hallucination is detected, \name then intervenes by adjusting the LM's hidden states such that the model will produce more factual outputs.
\name provides fresh insights that the inner workings of LMs can be revealed by their hidden states.
Practically, both its detection and mitigation models are lightweight, adding little inference overhead; \name proves a more efficient approach for mitigating hallucinations compared to many post-hoc alternatives.
We evaluate \name over LMs of different scales and model families (including Llama, Mistral, Qwen and Gemma),
across a variety of QA datasets from different domains.
Our results demonstrate the effectiveness of \name, achieving over 70\% \textit{preemptive} detection accuracy. On average, outputs generated by LMs with intervention are 34.4\% more factual compared to those without.
\end{abstract}

\section{Introduction}
Language models (LMs) hallucinate, a phenomenon where they produce nonfactual or even misleading outputs that often appear plausible \citep{10.1145/3571730, bang2023multitaskmultilingualmultimodalevaluation,xu2024hallucination,zhang2023siren,li2024dawn, huang2023surveyhallucinationlargelanguage, ye2023cognitivemiragereviewhallucinations}.
% \hao{i'm sure there are tons of more papers to cite}. \deema{added}
% by generating nonfactual outputs% However, a significant impediment to their broader adoption is the phenomenon of hallucinations, where LLMs generate misleading, incorrect, or nonsensical content without any intent to deceive %\cite{park2023ai}
% , which severely undermines their reliability in real-world applications \citep{10.1145/3571730, bang2023multitaskmultilingualmultimodalevaluation}.
%such as medical advice and news generation. \hao{can we connect to the experiments we did?} \deema{how?}, 
% \hao{i'd expand the following in to several sentences that briefly review how previous works address this} %addressed
Extensive efforts have been devoted to mitigating their hallucination issues \citep{min2023factscorefinegrainedatomicevaluation, manakul2023selfcheckgpt, rawte2023surveyhallucinationlargefoundation, zhou2021detectinghallucinatedcontentconditional}. 
These approaches are mostly \emph{reactive}, addressing hallucinations \textit{after} they occur, and
often require resampling new outputs \citep{li2023haluevallargescalehallucinationevaluation, manakul2023selfcheckgptzeroresourceblackboxhallucination}, substantially increasing the inference overhead. 
In addition, they often treat the LM as a black box, while relying on external LMs for detecting hallucinations,
missing the opportunity to gain deeper insights into the internal workings of these models.
% Therefore, they are extrinsic in nature and their performance is upper-bounded by the external LM used.

\begin{figure*}[th]
    \centering
    \includegraphics[width=\textwidth]{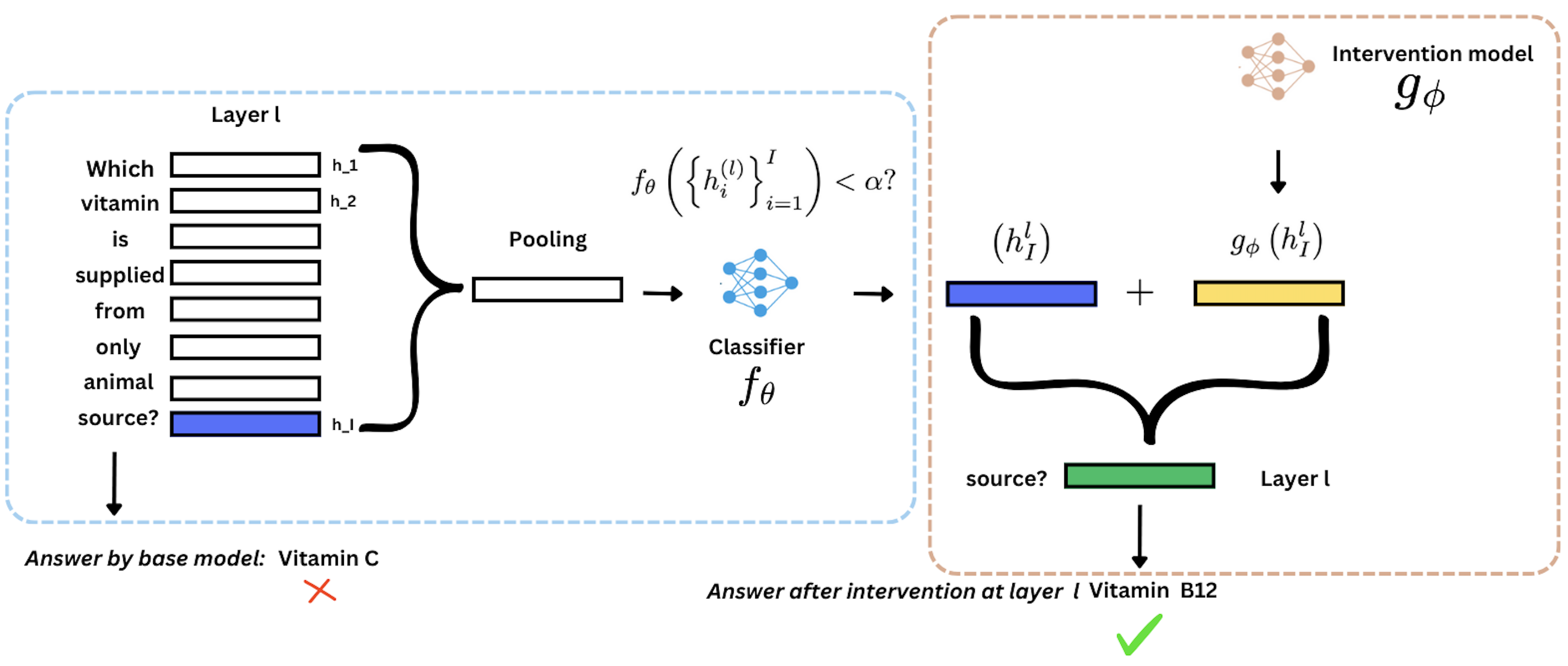}
    \caption{\name P \textit{at inference time}. 
    A demonstration of how preemptive detection and subsequent mitigation work. 
    As shown, at a layer $l$, the hidden states of only the prefix $I$ are aggregated and passed to the classifier \classifier (highlighted in the light blue box \S\ref{sec:detect}). Once hallucination is detected with classification probability \(< \alpha\), \intervention (highlighted in the light brown box \S\ref{sec:intervene}) intervenes and adjusts the last token \(\mathbf{h}^{(l)}_I\). This leads to a more factual output than before.
    }
    \label{fig:pipeline}
    % \vspace{-.5cm}
\end{figure*}

Recent findings by \citet{azaria2023internal} and \citet{burns2022discovering}  show that the LMs' representations can provide useful information about the factuality of their outputs.
\citet{marks2023geometry} observe that LMs' hidden states generated over factual and non-factual statements are linearly separable.
However, these studies have a relatively narrow focus, primarily addressing hallucination detection in a \emph{reactive} manner, 
and a more thorough investigation is needed.
% A recent approach to mitigating hallucinations involves utilizing activation engineering \citep{subramani2022extractinglatentsteeringvectors}, first applied to hallucination mitigation by \citet{duan2024llmsknowhallucinationempirical}. 
% \hao{this one should go to related works:\name builds on these findings and explores additional activation engineering techniques to intervene and mitigate hallucinations during inference time.}
% \neeraja{similar sentence already there.}
% The concept of utilizing activation engineering \citep{subramani2022extractinglatentsteeringvectors} to mitigate hallucination was initially demonstrated by \citep{duan2024llmsknowhallucinationempirical}. 
% \name extends these findings by exploring additional activation engineering mitigating hallucination techniques to intervene at inference time

The key hypothesis of this paper is that, the LMs' hidden states reveals valuable information about their internal working mechanisms, and provide signals that can be used to predict whether it will hallucinate \emph{before} decoding.
We propose \name to answer the following research question (RQ):
\emph{Can we \textbf{preemptively} predict and mitigate hallucinations with LMs' internal representations?}
\name learns a classifier that, taking the models' hidden states over the inputs, predicts whether the model \emph{is about to} hallucinate.
If a hallucination is detected, \name intervenes, by adjusting the LM's hidden states with a learned intervention model, and steering them towards producing more factual outputs (Figure~\ref{fig:pipeline}).

% \hao{i didn't get the reason for adding this text below:}\textcolor{olive}{The focus of the paper is addressing factuality hallucination, where hallucination is defined when a model outputs an incorrect answer to a given question. The benefit of preemptively detecting hallucinations is that it 1) Reduces computational costs by avoiding expensive post-hoc corrections or re-sampling strategies, 2) Reduces error propagation by addressing hallucinations at their source before they influence further decoding steps. 3) Builds trust in LLM applications by providing early-warning signals and proactive intervention mechanisms.}
%, at least for the question answering (QA) task.
% We focus on QA tasks 
% their controlled nature allows us to precisely identify when hallucinations occur.
Our controlled experiments answer the RQ in the positive.
We evaluate \name across four QA datasets from different domains: NQ-open (Wikipedia; \citealp{lee-etal-2019-latent}), MMLU (STEM exams; \citealp{mmlu}), MedMCQA (medical;~\citealp{pal2022medmcqa}), and GSM8K (Math; ~\citealp{cobbe2021trainingverifierssolvemath}).
Using the LMs' hidden states over the input questions,
\name can successfully predict whether or not the LMs will hallucinate over 70\% of the time, significantly outperforming a 50\% random baseline (\S\ref{sec:detect}).
% For all, \name successfully predicts whether or not the LMs will hallucinate over 70\% of the time, significantly outperforming a 50\% random baseline.
% This is achieved when the LMs have only seen the input questions and before decoding starts.
We observe consistent trends across base and fine-tuned LMs of different scales and families, including Llama2 (7B and 13B; \citealp{touvron2023llamaopenefficientfoundation}), Llama3 (8B and 70B) and Llama3.1 (8B and 70B) \citep{llama3} 
Mistral-7B \citep{mistral}, Gemma-7B \citep{gemma}, and Qwen2.5 (7B and 32B) \citealp{qwen2025qwen25technicalreport}.
% \hao{clarify these two settings below: when trained on all models/datases, \name can handle various models/tasks}

We further conduct cross-model and cross-dataset evaluations to assess the generalizability of \name. In the cross-model training, we train on the hidden states of multiple models. Whereas, in the cross-dataset, we train the classifier on multiple domains. The cross-model classifier achieves on average 71\% accuracy in preemptive hallucination detection, demonstrating its robustness across different models. Similarly, the cross-dataset classifier attains an average of 65\% accuracy.
%59\% 
Furthermore, \name's intervention model can effectively improve the LMs' factuality.
Using GPT-4o as a judge, which shows high agreement with human evaluations in our experiments, we find that on average, outputs generated by LMs with intervention are 34.4\% more factual than those produced without intervention (\S\ref{sec:intervene}).
% outputs generated after intervention are considered more factual up to 77\% of the time, whereas outputs without intervention are preferred at best 41\% of the time.  %\hao{check the numbers}. 
We also calculate the inference time overhead introduced by \name, incurring minimal average overhead of a 1.2\% increase in decoding time,
showing minimal impact on inference efficiency (\S\ref{time_eval_sec}).
\name reveals surprising  insights into existing LMs, and can potentially lead to more profound understanding of their internal working. 
All code, data, and checkpoints for reproducing our findings will be released.

\section{Related Work and Motivation}

% \hao{related work is usually not that important, unless it lays some background, which does not seem to be the case here.
% i'd consider moving it to the second to the last section
% }

% \mk{I suggest you combine mitigation + detection (all in one parahraph), 
% % and you need another section for guided decoding techniques,n % we are not touching decoding technique
% and one for interpreatbility of hidden states. }
% % You can use the related work section in my GRACE paper related work on guided decoding.}
% \hao{not sure we need the paragraph below} \deema{to state that we are not feeding the LM external contexual knowledge and only rely on its parametric knowledge}

% \textbf{Definitions.} 
\textbf{Hallucination Detection.}
We investigate hallucinations in language models that generate responses based solely on their parametric knowledge, similar to~\citet{azaria2023internal}. This contrasts with in-context generation scenarios where external knowledge sources are explicitly incorporated within the prompt. 
We focus on addressing factuality hallucinations, an important type of hallucinations as argued in ~\citet{huang2023surveyhallucinationlargelanguage}.

Existing research primarily focuses on post-processing methods applied after the inference process is completed and often utilizing external knowledge for verification \citep{manakul2023selfcheckgptzeroresourceblackboxhallucination, li2023haluevallargescalehallucinationevaluation, chern2023factoolfactualitydetectiongenerative}. For instance, CRITIC \citep{gou2024criticlargelanguagemodels} validates model outputs through tool interactions, and FACTSCORE \cite{min2023factscorefinegrainedatomicevaluation} breaks down generated content into atomic facts, assessing their accuracy by comparing them against reliable sources. \par
A recent line of research leverages the internal mechanics of LMs to detect hallucinations \citep{burns2024discoveringlatentknowledgelanguage, azaria2023internal, marks2023geometry}. 
\citet{meng2022locating} locates where factual associations are stored in GPT models. These studies have spurred further research into using LMs' internal representations in hallucination detection ~\citep{chen2024insidellmsinternalstates, chwang2024androidsknowtheyredreaming}.
For instance, MIND \cite{su2024unsupervisedrealtimehallucinationdetection} generates training data in unsupervised approach for training hidden states based hallucination detectors. \citet{duan2024llmsknowhallucinationempirical} conducts an experimental examination of the hidden states of LLMs when processing factual versus nonfactual responses. 
\textsc{FactCheckmate} demonstrates the effectiveness of preemptive hallucination detection, identifying warning signals several tokens \textbf{before} the hallucinations actually occur, via the language model's hidden states. 

%unlike existing works focused on analyzing the output vocabulary distribution (Chen et al., 2022), ours explores the LLM’s internal representation space, opening up exciting new possibilities for understanding LLM hallucination.
% no sampling needed///X
%
% \hao{this one should go to related works:}
\textbf{Hallucination Mitigation.}
For hallucination mitigation at inference time, existing works have explored self-correction and automated feedback approaches, where the language model is prompted to fix its generation flaws, with or without leveraging feedback from the model itself or some external knowledge source \citep{pan2023automaticallycorrectinglargelanguage, dhuliawala2023chainofverificationreduceshallucinationlarge, ji2023mitigatinghallucinationlargelanguage}.
A recent approach involves utilizing activation engineering \citep{subramani2022extractinglatentsteeringvectors,duan2024llmsknowhallucinationempirical, zhang2024truthxalleviatinghallucinationsediting}. 
\name builds on these findings and explores activation engineering techniques to preemptively intervene and mitigate hallucinations during inference time. It is also related to inference-time approaches that utilize a scoring function to steer the LM toward desired behavior \citep{pplm,khalifa2023grace}.

\textbf{Hidden States as Predictive Signals of Hallucination.}
Previous works primarily use hidden states as indicators of factuality after generation \citep{chuang2024doladecodingcontrastinglayers, zhang2024truthxalleviatinghallucinationsediting, li2024inferencetimeinterventionelicitingtruthful,orgad2024llmsknowshowintrinsic}.
\name instead asks: Can hidden states reveal early signals of hallucination before tokens are generated? If so, this would suggest that factuality cues are embedded in the model's internal mechanisms earlier than previously assumed.
\name demonstrates, for the first time, that hidden states offer early signals correlated with hallucination.
Its finding reveals that factuality cues are embedded within the model's internal mechanisms well before the output is generated.
This fresh insights allows \name to use the model's hidden states to anticipate when it is likely to hallucinate, rather than waiting for errors to surface in generated tokens.
This reduces the need for expensive post-hoc corrections and provides insight into how factual knowledge is internally encoded and accessed.
Our results (\S\ref{sec:detect}) suggest that hallucinations are not merely failures of token-level prediction but often emerge from systematic patterns in how models encode factual information during inference \citep{zou2023representationengineeringtopdownapproach}.
\section{Preemptive Hallucination Detection}\label{sec:detect}

This section focuses on \name's preemptive hallucination classifier (\S\ref{subsec:classifier}) and experimental results (\S\ref{subsec:classifier_results}).

\subsection{Preemptive Hallucination Detection with a Lightweight Classifier over Hidden States}\label{subsec:classifier}

\paragraph{Classifier.}\name learns a binary classifier \classifier to preemptively detect hallucinations.
Parameterized by a learned two-layer $\relu$-MLP followed by a sigmoid function, \classifier takes as input the LM's hidden states and outputs the probability that the LM \emph{will} hallucinate.
More specifically, let $\{\vh^{(l)}_i\}_{i=1}^{I}$ be a sequence of $I$ hidden states that the LM produces over the input question with $I$ tokens.
A $d$-dimensional vector $\vh^{(l)}_i$ denotes the output of the feedforward network (FFN) of the $l$-th transformer layer, at the $i$-th token.\footnote{
We utilize the outputs of the FFNs, following previous work by \citep{marks2023geometry, azaria2023internal}, as the FFN module is commonly regarded as a knowledge memory \cite{hernandez2024inspectingeditingknowledgerepresentations}
}.
%, which lead to better performance than alternatives (e.g., attention layers) in our preliminary experiments.\hao{i added this. please check whether this is accurate}
% \deema{there is no clear answer to this in the literature and we will add this experiment after submission}

The classifier \classifier takes as  input the pooled values over $\{\vh^{(l)}_i\}_{i=1}^{I}$ and produces a scalar between 0 and 1 indicating the probability that the LM will hallucinate in its response to the input: $f_\theta (\{\vh^{(l)}_i\}_{i=1}^{I})=$
% \begin{align*}
% f_\theta \Big(\{\vh^{(l)}_i\}_{i=1}^{I}\Big) = 
% \sigmoid\Bigg(\operatorname{ReLU-MLP}\Big(\frac{1}{I}\sum_{i=1}^{I}\vh^{(l)}_i\Big)\Bigg)
% \end{align*}
\begin{align*}
\sigmoid\Bigg(\operatorname{ReLU-MLP}\Big(\mathcal{A}\Big(\{\vh^{(l)}_i\}_{i=1}^{I}\Big)\Big)\Bigg)
\end{align*}
where $\mathcal{A}$ represents the pooling function, which can be the mean, max, or selecting the last token.
$l$ is empirically determined based on validation performance, and can vary by the LMs and datasets.
In general, $l$ tends to be the middle to last layers. More details about the best empirical layer for each LM can be found in Appendix~\ref{Classification_analysis}.

We train a separate classifier tailored to each LM.
% \footnote{
% Our preliminary experiments indicate that the hallucination classifier performs suboptimally when applied to hidden states generated by a model different from the one it was trained on. This aligns with the observation that different models encode hidden state spaces differently. \textcolor{red}{However, training on a diverse model pool improves generalization. cite}
% }
We consider LMs from different families of different scales, including Llama2 (7B and 13B; \citealp{touvron2023llamaopenefficientfoundation}), Llama3 (8B and 70B) and 3.1-8B (8B and 70B;~\citealp{llama3}), 
Mistral-7B \citep{mistral}, Gemma-7B \citep{gemma}, and Qwen2.5 (7B and 32B; \citealp{qwen2025qwen25technicalreport}) for base and fine-tuned versions.

% \hao{move to experiments: Our experiments also explore taking the elementwise max over hidden states, or taking the last one as the input to \classifier, and find they slightly underperform taking the average as explained in section \ref{Classification_analysis}.}
% \mk{the last 3 sentences might be removed. Sufficies to mention them in the experimental setting section.} 

\paragraph{Data collection and training.}
\label{subsec:data_collection}
In order to train \classifier, we need to collect paired data consisting of the LMs' hidden states and binary labels indicating whether it will hallucinate.
We construct the training data on four QA datasets from different domains: NQ-open (Wikipedia; \citealp{lee-etal-2019-latent}), MMLU (STEM; \citealp{mmlu}), MedMCQA (medical entrance exam; \citealp{pal2022medmcqa}), and GSM8K (Math; ~\citealp{cobbe2021trainingverifierssolvemath}).
NQ-open is a QA dataset and contains question and answer pairs.
MMLU and MedMCQA are multiple choice datasets, pairing each question with multiple options. 
We convert MMLU and MedMCQA into a QA dataset by pairing each input question with the gold answer. GSM8K consists of grade school math problems, where each problem takes between 2 and 8 steps, we convert GSM8K into a QA dataset by pairing each problem with the final answer.

\begin{figure}[ht]
    \centering
    \includegraphics[width=\linewidth]{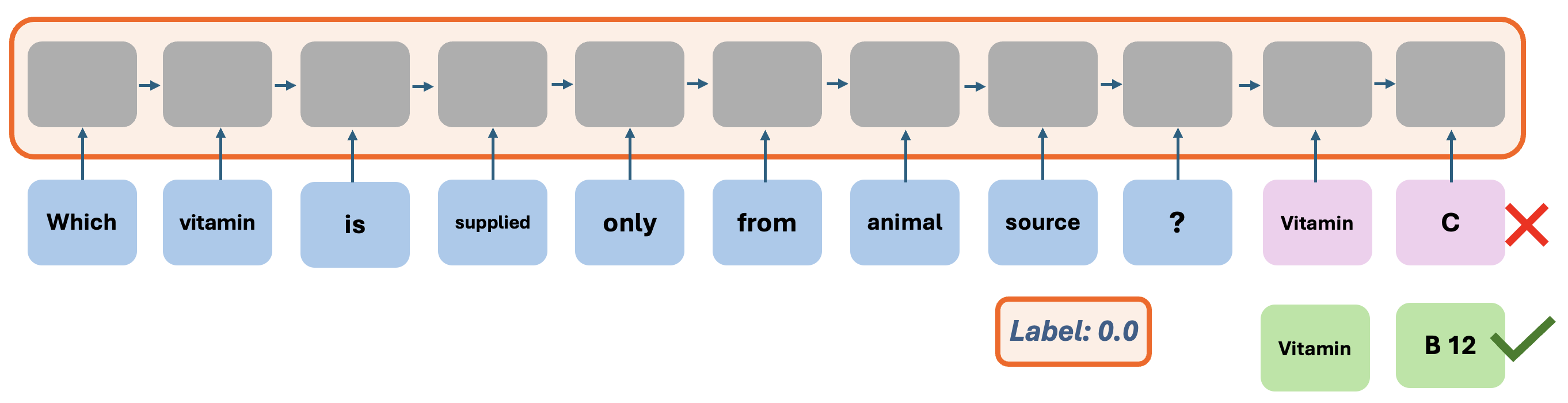}
    \caption{Example of the data collection process. We capture the hidden states over the input and output then label them based on EM with the gold outputs. (\S\ref{subsec:data_collection}).}
    \label{fig:data_collection_pipeline}
    % \vspace{-.2cm}
\end{figure}

To collect the training data for LM $M$, we prompt $M$ with 
few-shot demonstrations followed by a question, and then collect its hidden states over the inputs and the outputs.
$M$'s output answers are checked against gold ones with the exact match (EM), following standard practice~\citep{eval-harness}.
If the model's output is wrong, its associated hidden states are labeled nonfactual and vice versa, as shown in Figure~\ref{fig:data_collection_pipeline}.
After producing hidden state and label pairs, we subsample the data to obtain balanced training data containing roughly the same amount of positive (factual) and negative (nonfactual) pairs. 
% \textcolor{olive}{We get a balanced corpus to ensure that the model is not biased towards one particular label.}
In order to compare across different LMs, we create a shared test split across all LMs. Each LM have different training/validation splits.
Table~\ref{tab:data_sizes} in Appendix \ref{app:datasetsize} summarizes the statistics of the datasets.
\classifier is trained with a cross-entropy loss on the input's hidden state and label pairs.
Early stopping based on the validation accuracy is used. Other training details are explained in \ref{Classification_analysis}.
% \begin{figure}
%     \centering
%     \begin{subfigure}[b]{0.5\textwidth}
%         \centering
%         \includegraphics[width=0.7\textwidth]{figures/data_collection.png}
%         \caption{Data Collection Pipeline}
%         \label{fig:data_collection}
%     \end{subfigure}
%     \caption{Overview of the data collection process. This figure illustrates the pipeline for collecting, processing, and annotating data used in our experiments (\S\ref{subsec:data_collection}).}
%     \label{fig:data_collection_pipeline}
% \end{figure}

We choose to focus on short-answer QA tasks because they allow for unambiguous evaluation using exact match (EM) and controlled experiments. Besides, the short answers allow us to identify exactly where hallucinations begin, as our approach involves preemptively analyzing hidden states before hallucinations occur. 
More concretely, for a non-factual output, we can use the first token in the wrong answer as the starting point of the hallucination.

% This setup ensures clear ground truth labels, making it easier to pinpoint hallucination onset and simulate using LMs to generate content in practice.

% \begin{table*}[htbp]
%     \centering
%     \begin{tabular}{@{} lrrrr@{}}
%         \toprule
%         \textbf{Dataset} & \textbf{Total Size} & \textbf{Train (70\%)} & \textbf{Validation (15\%)} & \textbf{Test (15\%)} \\
%         \midrule
%         NQ-Open~\citep{lee-etal-2019-latent} & 6,666 & 4,666 & 1,000 & 1,000 \\
%         MMLU~\citep{mmlu} & 3,182 & 2,228 & 477 & 477 \\
%         MedMCQA~\citep{pal2022medmcqa} & 3,953 & 2,767 & 593 & 593 \\
%         GSM8K~\citep{cobbe2021trainingverifierssolvemath} & 1,040 & 728 & 156 & 156 \\
%         \bottomrule
%     \end{tabular}
%     \caption{Dataset splits and sizes for training the hallucination classifier \classifier over the LMs' hidden states (\S\ref{subsec:classifier}). }
%     \label{tab:data_sizes}
%     \vspace{-.3cm}
% \end{table*}

\subsection{Experiments}\label{subsec:classifier_results}

\begin{figure}[ht]
    \centering
    \includegraphics[width=0.48\textwidth]{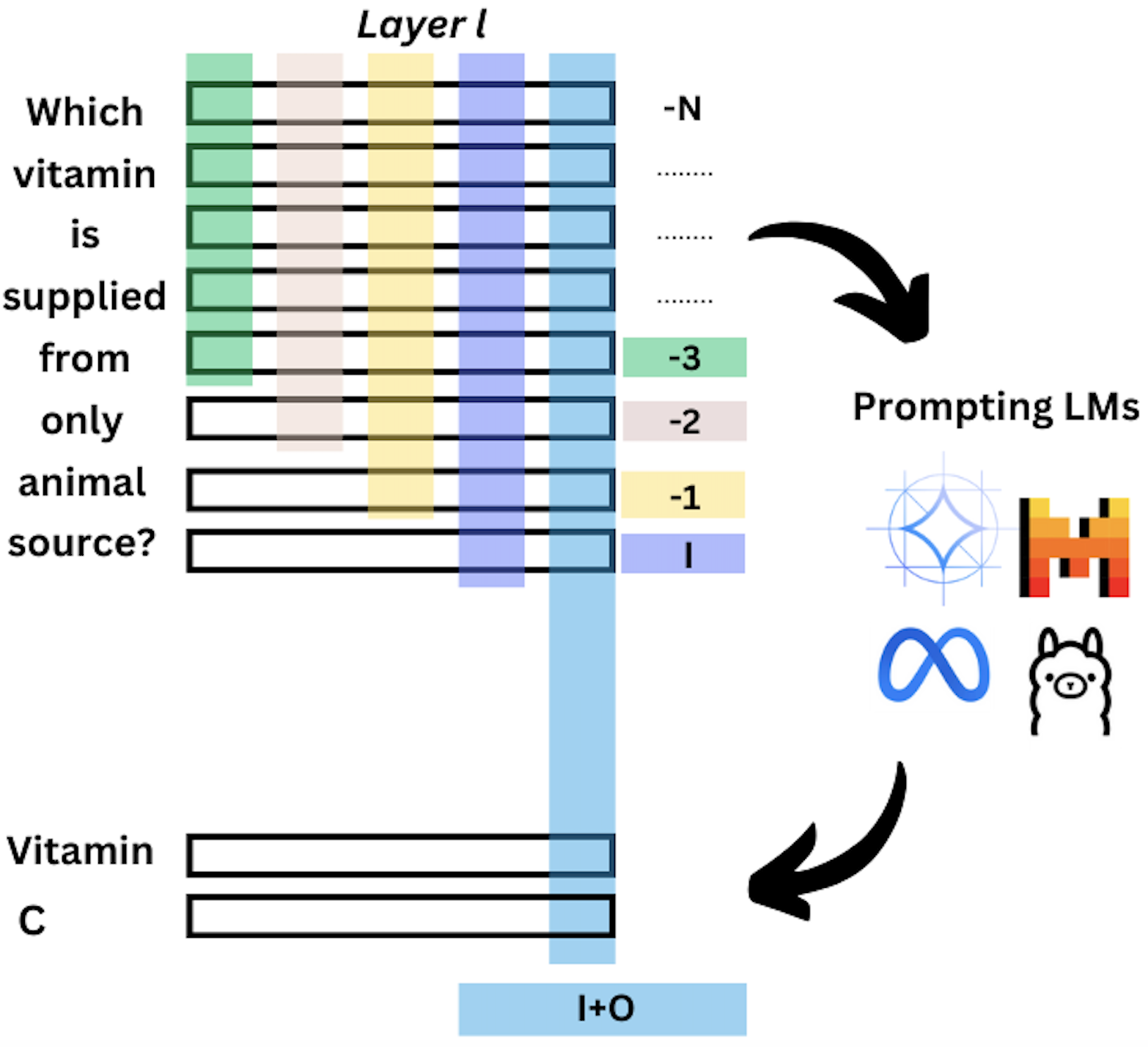}
    \caption{An illustration of different settings used in the experiment. \textbf{(Input + Output)} are the hidden states of both the input and output. The subsequent hidden states in the experiment are taken by using only the input or by dropping the last $n$ tokens from the prefix.
    }
    % \vspace{-.5cm}
    \label{fig:prefix-explanation}
\end{figure}

% \hao{this section should be concise. we only need to talk about table 2. be concise and focus on the highlights. talk about numbers and conclusions.}

% \hao{points to cover:
% (1) I performs way above a 50\% random guess baseline. indicating useful signal in the hidden states can be used to predict hallucinations \deema{addressed}
% (2) sometimes the prefix underperforms I, sometimes it is close \deema{addressed}
% (3) unsurprisingli, I underperforms a I+O reactive classifier. \deema{addressed}
% (4) hint the word embedding experiments---we are indeed classifying hidden states, instead of questions \deema{addressed below}
% }
%%%%%%%%%%%%%%%%%%%%%%%%%%%%%%%%%%%%%%
% \hao{suggest moving the content explaining the settings to a dedicated paragraph here, including the baselines and I-n etc.}

 % >{\centering\arraybackslash\columncolor[HTML]{A1D1F0}}m{2em}
 %    >{\centering\arraybackslash\columncolor[HTML]{B0BBFA}}m{2em}
 %    >{\centering\arraybackslash\columncolor[HTML]{FCEFB7}}m{2em}
 %    >{\centering\arraybackslash\columncolor[HTML]{F4E9E4}}m{2em}
 %    >{\centering\arraybackslash\columncolor[HTML]{9BDEB8}}m{2em}
    
\paragraph{Setting.}
We evaluate \classifier with the settings below: %on different inputs. 
% We use the best performing layer for every model used.
\begin{itemize}[noitemsep,topsep=0pt,parsep=0pt,partopsep=0pt]
    \item \colorbox[HTML]{B0BBFA}{\bf I:} our \emph{preemptive} classifier. It takes the LMs' hidden states produced over the \term{input questions only}.
    \item \colorbox[HTML]{FCEFB7}{\bf I-$n$} takes it even further, restricting \name's access to only prefixes of the input questions that exclude the last $n$ tokens.
See Figure \ref{fig:prefix-explanation} for illustrative diagrams.
 \item\colorbox[HTML]{A1D1F0}{\bf I+O} is a reactive setting, and is not to be compared to \name because it leverages additional information from a concatenation of both the input questions and the models' outputs and is expected to perform better.
Rather, it  serves as an approximation of the ceiling performance.
\end{itemize}

\paragraph{Results.}
Table~\ref{tab:classification} shows the hallucination detection test accuracy. We see that across all sizes (7B to 70B), various families (Llama2, Llama3, Mistral, Gemma, Qwen), various types (base, chat, instruct) \colorbox[HTML]{B0BBFA}{\bf I} achieves competitive performance, sometimes even comparable to \colorbox[HTML]{A1D1F0}{\bf I+O} (e.g., only a 1\% gap with Llama2-13b on MMLU)
% \textcolor{red}{(e.g., only a 2.9\% gap with Llama3-8B-Instruct on NQ)}. 
This means that the hidden states are capable of predicting the hallucination just by looking at the question, i.e even before the model outputs any incorrect answer.
% (e.g., only a 2.4\% gap with LLama3-70B on NQ).

% is comparable to the upper bound which is the accuracy of classifier across the hidden states of both the input and output. 
% Throughout all {\bf I} settings across all LMs and datasets, \classifier achieves well above the 50\% random guess baselines.
This confirms that LMs' hidden states provide useful signal for predicting their hallucinations \textit{preemptively}.
In some cases, using a prefix of \colorbox[HTML]{FCEFB7}{\bf I-$n$} underperforms \colorbox[HTML]{B0BBFA}{\bf I}, while for others their performance is comparable.
These results suggest that \classifier can often predict whether the LM is likely to hallucinate before it even finishes processing the input questions.
\begin{table*}[!t]
\centering
\begin{adjustbox}{max width=\textwidth}
\renewcommand{\arraystretch}{1.2} % Adjust row spacing
\setlength{\tabcolsep}{4pt}       % Adjust column padding
\begin{tabular}{
    l
    >{\centering\arraybackslash\columncolor[HTML]{A1D1F0}}m{2em}
    >{\centering\arraybackslash\columncolor[HTML]{B0BBFA}}m{2em}
    >{\centering\arraybackslash\columncolor[HTML]{FCEFB7}}m{2em}
    >{\centering\arraybackslash\columncolor[HTML]{F4E9E4}}m{2em}
    >{\centering\arraybackslash\columncolor[HTML]{9BDEB8}}m{2em}
    m{0.01em}
    >{\centering\arraybackslash\columncolor[HTML]{A1D1F0}}m{2em}
    >{\centering\arraybackslash\columncolor[HTML]{B0BBFA}}m{2em}
    >{\centering\arraybackslash\columncolor[HTML]{FCEFB7}}m{2em}
    >{\centering\arraybackslash\columncolor[HTML]{F4E9E4}}m{2em}
    >{\centering\arraybackslash\columncolor[HTML]{9BDEB8}}m{2em}
    m{0.01em}
    >{\centering\arraybackslash\columncolor[HTML]{A1D1F0}}m{2em}
    >{\centering\arraybackslash\columncolor[HTML]{B0BBFA}}m{2em}
    >{\centering\arraybackslash\columncolor[HTML]{FCEFB7}}m{2em}
    >{\centering\arraybackslash\columncolor[HTML]{F4E9E4}}m{2em}
    >{\centering\arraybackslash\columncolor[HTML]{9BDEB8}}m{2em}
    m{0.01em}
    >{\centering\arraybackslash\columncolor[HTML]{A1D1F0}}m{2em}
    >{\centering\arraybackslash\columncolor[HTML]{B0BBFA}}m{2em}
    >{\centering\arraybackslash\columncolor[HTML]{FCEFB7}}m{2em}
    >{\centering\arraybackslash\columncolor[HTML]{F4E9E4}}m{2em}
    >{\centering\arraybackslash\columncolor[HTML]{9BDEB8}}m{2em}
}
\toprule
& \multicolumn{5}{c}{\textbf{NQ}} 
&& \multicolumn{5}{c}{\textbf{MMLU}} 
&& \multicolumn{5}{c}{\textbf{MedMCQA}}
&& \multicolumn{5}{c}{\textbf{GSM8K}} \\

\cmidrule(lr){2-6} \cmidrule(lr){8-12} \cmidrule(lr){14-18} \cmidrule(lr){20-24}

& \cellcolor{white} \textbf{  } & \multicolumn{4}{c}{\bf Preemptive (ours)} 
&& \cellcolor{white} \textbf{  } & \multicolumn{4}{c}{\bf Preemptive (ours)} 
&& \cellcolor{white} \textbf{  } & \multicolumn{4}{c}{\bf Preemptive (ours)}
&& \cellcolor{white} \textbf{  } & \multicolumn{4}{c}{\bf Preemptive (ours)} \\

\cmidrule(lr){3-6} \cmidrule(lr){9-12} \cmidrule(lr){15-18} \cmidrule(lr){21-24}

\textbf{LM} 
& \textbf{I+O} & \textbf{I} & \textbf{-1} & \textbf{-2} & \textbf{-3}
&& \textbf{I+O} & \textbf{I} & \textbf{-1} & \textbf{-2} & \textbf{-3}
&& \textbf{I+O} & \textbf{I} & \textbf{-1} & \textbf{-2} & \textbf{-3} 
&& \textbf{I+O} & \textbf{I} & \textbf{-1} & \textbf{-2} & \textbf{-3} \\

\midrule
Llama2-7B   & 72.8 & 71.8 & 71.1 & 68.1 & 65.7 %
&& 91.7 & 91.9 & 91.7 & 91.7 & 91.7 %
&& 77.0 & 72.9 & 72.9 & 72.9 & 74.5 %%
&& 65.8 & 66.0 & 66.0 & 63.5 & 63.5\\

Llama2-13B      & 74.4 & 72.0 & 70.6 & 71.4 & 69.7 %
&& 94.0 & 93.0 & 84.1 & 92.7 & 85.7 %
&& 76.0 & 78.3 & 78.6 & 78.3 & 74.2 %%
&& 68.4 & 69.1 & 68.4 & 66.8 & 63.8\\

\midrule
Llama3-8B 
& 74.9 & 70.2 & 68.5 & 66.8 & 66.8 %
&& 93.8  & 94.0 & 87.5 & 87.1 & 77.3 %
&& 77.1 & 76.3 & 74.3 & 71.2 & 67.3 %%
&& 71.3 &  72.5 & 72.9 & 71.3 & 66.2 \\

Llama3.1-8B  & 74.3 & 73.1 & 70.9 & 68.9 & 68.1 && %
94.5 & 92.3 & 86.3 & 80.0 & 78.0 %
&& 78.4 & 76.2 & 74.9 & 73.6 & 69.4 %%
&& 72.3 & 69.1 & 61.2 & 60.2 & 60.6\\

% Llama3-70B      & 71.5 & 69.1 & 66.4 & 65.0 & 64.3 && -    & -    & -    & -    & -    
% && -    & -    & -    & -    & - 
% && 78.0 & 77.2 & 76.8 & 76.8 & 76.5 \\

\midrule
Mistral-7B  & 73.3 & 72.5 & 71.4 & 71.1 & 70.3 %
&& 93.2  & 90.2  & 83.0  & 82.5 & 82.8 %
&& 77.9 & 75.4 & 75.2 & 73.9 & 72.8 %%
&& 69.4 & 70.0 & 70.0 & 71.8 & 71.8 \\

\midrule
Gemma-7B   & 80.2 & 74.5 & 74.4 & 74.2 & 73.9 %running
&& 92.2 & 96.9 & 91.3 & 81.5 & 89.6 %
&& 77.0 & 77.5 & 74.7 & 75.2 & 75.2  %%
&& 70.9 & 67.4 & 67.0 & 67.4 & 67.8\\

\midrule
Qwen2.5-7B     & 76.1 & 74.5 & 72.7 & 71.2 & 69.2 %
&& 94.3 & 94.0 & 71.3 & 85.4 & 74.4 %
&& 78.9 & 76.6 & 76.9 & 75.7 & 74.9 %%
&& 67.0 & 67.2 & 67.2 & 67.2 & 67.0 \\

% Qwen2.5-32B     & - & - & - & - & -
% && 92.7 & 90.8 & 86.9 & 87.3 & 74.4 %
% && -  & - & - & - & - 
% && 79.8 & 76.3 & 75.9 & 75.5 & 75.9\\

\midrule
Llama2-7B-chat & 76.2 & 74.5 & 67.5 & 69.4 & 69.2 %
&& 94.8 & 90.9 & 79.1 & 83.3 & 83.0 %
&& 81.1  & 79.3 & 79.3 & 79.0 & 79.3 %%
&& 72.3  & 73.6 & 72.2 & 72.9 & 72.2 \\

Llama2-13B-chat & 74.8 & 72.4 & 70.7 & 70.9 & 68.5 %
&& 93.8 & 93.9 & 80.1 & 78.6 & 92.1 %
&& 81.3  & 73.3 & 70.8 & 70.0 & 71.9 
&& 72.3 & 71.9 & 71.9 & 71.9 & 71.9 \\

\midrule
Llama3-8B-Instruct & 81.5 & 78.6 & 77.2 & 76.4 & 75.0 
&& 93.8 & 95.6 & 87.4 & 85.0 & 79.5  %
&& 81.4 & 77.3 & 72.2 & 71.1 & 68.8 %%
&& 74.7 & 74.3 & 74.3 & 74.3 & 74.3\\

Llama3.1-8B-Instruct & 83.3 & 74.5 & 71.3 & 70.9 & 66.7
&& 93.1 & 91.8 & 86.4 & 85.4 & 80.1 %
&& 81.7  & 78.8 & 76.5 & 71.7 & 70.1 %%
&& 76.2  & 78.4 & 78.0 & 78.0 & 78.4 \\

Llama3-70B-Instruct & 81.0 & 77.1 & 73.3 & 69.6 & 65.9
&& 87.6  & 79.6 & 76.5 & 76.4 & 73.4 
&& 74.7  & 67.6 & 64.5 & 63.6 & 61.0
&& 82.7  & 78.8 & 72.5 & 71.3 & 69.4 \\

% Llama3.1-70B-Instruct & -  & - & - & - & - 
% && -  & - & - & - & - 
% && -  & - & - & - & - 
% && 85.3  & 81.0 & 77.6 & 70.9 & 77.2 \\

\midrule
Mistral-7B-Instruct & 75.1 & 74.2 & 70.9 & 66.9 & 65.8 %
&& 94.3 & 93.9 & 89.7 & 70.7 & 90.8 %
&& 76.2 & 75.9 & 75.6 & 76.1 & 75.3 %%
&& 71.5 & 72.9 & 72.9 & 72.9 & 72.5 \\

\midrule
Gemma-7B-it     & 83.2 & 75.4 & 73.5 & 71.5 & 68.7 
&& 90.2 & 94.0 & 78.9 & 78.8 & 84.3 %
&& 77.0 & 76.2 & 76.2 & 75.0 & 75.2 %%
&& 74.7 & 75.5 & 61.1 & 64.6 & 70.2\\

\midrule
Qwen2.5-7B-Instruct & 74.4 & 72.6 & 70.2 & 69.9 &  66.6 %
&& 92.5 & 94.5 & 85.9 & 87.8 & 86.2 %
&& 82.0 & 77.5 & 75.7 & 73.9 & 72.3 %%
&& 76.4 & 80.4 & 74.8 & 67.4 & 67.8\\

% Qwen2.5-32B-Instruct  &. & . & . & . & .
% && 92.7 & 92.1 & 83.7 & 86.2 & 83.8 %
% && . & . & . & . & .
% && 72.3 & 72.5 & 72.5 & 72.5 & 72.5\\

\bottomrule
\end{tabular}
\end{adjustbox}
\caption{
Hallucination detection test accuracy. 
\colorbox[HTML]{A1D1F0}{\bf I+O} indicates a ``reactive'' baseline that classifies the LMs' hidden states produced over both input questions and output answers,
while \colorbox[HTML]{B0BBFA}{\bf I} preemptively classifies hallucinations based on the hidden states over only the inputs.
\colorbox[HTML]{FCEFB7}{\bf $-n$} indicates that the classifier only sees a prefix of the input excluding the last $n$ tokens.}
\label{tab:classification}
% \vspace{-.35cm}
\end{table*}

%%%%%%%%%%%%%%%%%%%%%^^^^^^^^^^^^^^^

% \begin{table*}
% \begin{adjustbox}{max width=\textwidth}
% \begin{tabular}{@{} lccccc @{}}
% \toprule
% & \multicolumn{5}{c}{\textbf{NQ}} \\
% \cmidrule{2-6}
% & \bf I+O & \bf I  & \bf -1 & \bf -2 &\bf  -3 \\
% \midrule

% Llama-3.1-8B-Instruct & 0.833 & 0.7454 & 0.7131 & 0.7094 & 0.6668 \\
% Gemma-7b-it & 0.832 & 0.7536 & 0.7349 & 0.7146 & 0.6870 \\
% Llama2-7b-chat-hf & 0.833 & 0.7406 & 0.6812 & 0.6557 & 0.6255 \\
% Llama2-13b-hf & 0.818 & 0.7641 & 0.7185 & 0.7216 & 0.7133 \\
% Llama3-70B & 0.715 & 0.6906 & 0.6641 & 0.6495 & 0.6432 \\
% Llama3-70B-Instruct & 0.8104 & 0.7708 & 0.7333 & 0.6964 & 0.6594 \\
% Llama3-8B-Instruct & 0.815 & 0.7857 & 0.7716 & 0.7638 & 0.7499 \\

% \bottomrule
% \end{tabular}
% \end{adjustbox}
% \caption{}
% \end{table*}

\paragraph{Cross-Model and Cross-Dataset Generalization.}
Results are shown in Table~\ref{tab:multi-model}.
For the  cross-model setting, \classifier is trained on the input hidden states of Llama2-7B, Llama-3.1-8B, and Mistral-7B. 
These LMs were chosen to ensure consistent embedding sizes.
Similar to the previous experiments Table~\ref{tab:classification} , the accuracy of using only the input hidden states (\colorbox[HTML]{B0BBFA}{\bf I}) remains comparable to using both input and output hidden states (\colorbox[HTML]{A1D1F0}{\bf I+O}) across all three test sets.
However, it struggles with model transferability, showing limited effectiveness when applied to unseen models (Table~\ref{tab:multi-model:ood} in Appendix~\ref{Generalization appendix})

For cross-dataset generalization, we train  \classifier on a combined dataset spanning NQ-open, MMLU, and MedMCQA. As seen in Table~\ref{tab:multi-data}, the classifier maintains consistent accuracy for \colorbox[HTML]{B0BBFA}{\bf I} across all three datasets. However, it performs reasonably well in generalizing to out-of-domain datasets.(Table~\ref{tab:gen-ood} in Appendix~\ref{Generalization appendix}).
In summary, training on diverse datasets and models provides an appealing and practical way for \name to generalize to various tasks and architectures.

\begin{table}[ht!]
\centering
\adjustbox{max width=\linewidth}{%
    \begin{tabular}{{@{}
    l
    >{\centering\arraybackslash\columncolor[HTML]{A1D1F0}}m{3em}
    >{\centering\arraybackslash\columncolor[HTML]{B0BBFA}}m{3em}
    >{\centering\arraybackslash\columncolor[HTML]{FCEFB7}}m{3em}
    >{\centering\arraybackslash\columncolor[HTML]{F4E9E4}}m{3em}
    >{\centering\arraybackslash\columncolor[HTML]{9BDEB8}}m{3em}
}}
        \toprule
        \textbf{Model} & \textbf{I + O} & \textbf{I} & \textbf{I - 1} & \textbf{I - 2} & \textbf{I - 3} \\
        \midrule
        Llama 2 7B   & 78.3 & 69.2 & 62.8 & 61.6 & 62.7 \\
        Llama 3.1 8B & 78.6 & 68.6 & 64.2 & 61.7 & 62.0 \\
        Mistral V0.3 7B & 80.4 & 76.4 & 76.0 & 75.8 & 75.7 \\
        \bottomrule
    \end{tabular}}
    \caption{Test accuracy results of different models on various test datasets. The models were trained on Llama-2-7B, Llama-3.1-8B, and Mistral-V0.3-7B on the NQ-open dataset.}
    \label{tab:multi-model}
\end{table}
% \subsection{Generalization across data}
% Similarly, we evaluate our classifier by training on multiple datasets. The following table shows the results of that. 
\begin{table}[ht!]
\centering
\adjustbox{max width=\linewidth}{%
    \begin{tabular}{{@{}
    l
    >{\centering\arraybackslash\columncolor[HTML]{A1D1F0}}m{3em}
    >{\centering\arraybackslash\columncolor[HTML]{B0BBFA}}m{3em}
    >{\centering\arraybackslash\columncolor[HTML]{FCEFB7}}m{3em}
    >{\centering\arraybackslash\columncolor[HTML]{F4E9E4}}m{3em}
    >{\centering\arraybackslash\columncolor[HTML]{9BDEB8}}m{3em}
}}
        \toprule
        \textbf{Test Data} & \textbf{I + O} & \textbf{I} & \textbf{I - 1} & \textbf{I - 2} & \textbf{I - 3} \\
        \midrule
        NQ-Open  & 79.2 & 72.9 & 68.0 & 68.1 & 68.9 \\
        MMLU      & 82.1 & 69.1 & 69.4 & 68.9 & 67.3 \\
        MedMCQA   & 75.4 & 59.3 & 54.2 & 54.6 & 54.6 \\
        \bottomrule
    \end{tabular}}
    \caption{Test accuracy results of Llama2-7b-hf after training a multi-data classifier on different test datasets. The model was trained on MMLU, NQ-Open, and MedMCQA.}
    \label{tab:multi-data}
\end{table}

% \subsection{Generalization across models}

\section{\name Preemptive Hallucination Mitigation}\label{sec:intervene}
% \hao{
% not done
% i propose the following structure.
% section 2 is about hallucination detection classifier:
% \begin{itemize}
% \item[2.1]: describe the classifier
% \item[2.2]: data describe how we collect the data to train the classifier.
% \item[2.3]: experimental details and discussion
% \end{itemize}
% section 3 is about intervention model:
% \begin{itemize}
% \item[2.1]: describe the classifier
% \item[2.2]: data describe how we collect the data to train the classifier.
% \item[2.3]: experimental details and discussion
% \end{itemize}
% section 4: remaining experiments and analysis.
% }
% This section we discuss the second part of the \name framework, hallucination mitigation using lightweight intervention models, and cover the experimental setup and results.
% \hao{this paragraph fits better in the intro:} % addressed

This section focuses on using \name to preemptively mitigate hallucinations,
including its intervention model (\S\ref{subsec:intervene}) and experiments (\S\ref{subsec:intervene_results}).

% the hallucination mitigation model of \name and its experimental results. 
% In this section, we present \name approach to preemptively mitigating hallucinations by adjusting the LM's hidden states when a potential hallucination is detected, thereby steering the model toward generating more factual outputs.
% \hao{do we need a dedicated paragraph to talk about data collection, like we did in the prev section?}
% \deema{addressed}

% These target hidden states are paired with their corresponding inputs $\vh_{N}^{(l)}$ to train \intervention. 

\subsection{The Intervention Model}\label{subsec:intervene}
\label{sec:factual_hidden_states_adjustment}
When \classifier detects that LM $M$ is about to hallucinate, \name relies on an \term{intervention model} \intervention to mitigate hallucinations preemptively.
Conditioning on $\vh_{I}^{(l)}$, 
\intervention generates a $d$-dimensional vector and adds it to $\vh_{I}^{(l)}$, \emph{before} the LM decodes.
\begin{equation}
\widetilde{\vh}_{I}^{(l)} = \vh_{I}^{(l)}  + \vg_\phi\Big(\vh_{I}^{(l)}\Big)
\end{equation} 
$\widetilde{\vh}_{I}^{(l)}$ is then used in place of $\vh_{I}^{(l)}$ for onward LM decoding.
In inference, the intervention is applied at the last hidden state of the input $\vh_{I}^{(l)}$, as it aligns with the natural progression of decoding and targets the point where hallucinations are most likely to arise.

Intuitively, \intervention is supposed to steer the LM's hidden state towards a ``target hidden state'' $\vh^{*(l)}$,
which is more likely to lead to a factual output.

We explore a deterministic and a stochastic \intervention:
\begin{itemize}[noitemsep,topsep=0pt,parsep=0pt,partopsep=0pt]
\item \textbf{The deterministic \intervention} is a three-layer ReLU-MLP. It trains by minimizing the mean squared error (MSE) between the adjusted hidden state \( \widetilde{\vh}_I^{(l)} \) and the target one $\vh^{*(l)}$.
\item 
\textbf{The stochastic \intervention} treats the adjustment vector as a random variable of multivariate Gaussian.
It applies a reparameterization trick: $\vg_\phi(\vh_{I}^{(l)})=\boldsymbol{\mu}(\vh_I^{(l)}) + \boldsymbol{\epsilon}\odot \boldsymbol{\sigma}(\vh_I^{(l)})$ for training. 
Two three-layer ReLU-MLPs are used to for $\boldsymbol{\mu}$ and $\boldsymbol{\sigma}$, with the first two layer shared.
Its training objective remains the same MSE loss.
One benefit of the stochastic \intervention is allowing for sampling the adjustment vectors during inference, which we explore in the experiments.
\end{itemize}

\paragraph{Data collection and training.} 
\intervention is trained on pairs of $\vh_{N}^{(l)}$, last token over the concatenation of the input and the output of $N$ sequence length, and $\vh^{*(l)}$.
When the LM answers the question correctly, no further modification is needed and $\vh^{*(l)}=\vh_{N}^{(l)}$.
However, when the model answers the question incorrectly, we set the $\vh^{*(l)}$ to the LM's final hidden state \emph{over the input prompt followed by the gold answer.}
We construct the training data on two QA datasets: NQ-open (Wikipedia; \citealp{lee-etal-2019-latent}), and MedMCQA (medical entrance exam; \citealp{pal2022medmcqa}).
Other training details are explained in \ref{Classification_analysis}.

\subsection{Experiments}\label{subsec:intervene_results}

% % \subsubsection{Hidden States Intervention Experiment}
% \definecolor{orange1}{HTML}{FFA500}
% \definecolor{blue1}{HTML}{4682B4}
% \definecolor{green1}{HTML}{9ACD32}
% \begin{figure}
%     \centering
%     \begin{minipage}{0.5\textwidth}
%         % \centering
%         \begin{tabular}{lll}
%         \textcolor{orange1}{\rule{6mm}{2mm}} Intervened wins & \textcolor{green1}{\rule{6mm}{2mm}} Ties & \textcolor{blue1}{\rule{6mm}{2mm}} Base wins
        
%         \end{tabular}
%     \end{minipage}
%     % \includegraphics[width=0.3\linewidth]{figures/legend.png}
%     \includegraphics[width=1.136\linewidth]{figures/intervention_3.png}
%     \caption{\name Intervention models Results}
%     \label{fig:intervention_results}
% \end{figure}

% \begin{figure}
%     \centering
%     \begin{minipage}{0.5\textwidth}
%         % \centering
%         \begin{tabular}{lll}
%         \textcolor{orange1}{\rule{6mm}{2mm}} Intervened wins & \textcolor{green1}{\rule{6mm}{2mm}} Ties & \textcolor{blue1}{\rule{6mm}{2mm}} Base wins
        
%         \end{tabular}
%     \end{minipage}
%     % \includegraphics[width=0.3\linewidth]{figures/legend.png}
%     \includegraphics[width=1.1\linewidth]{figures/comparisonwithpca.png}
%     \caption{Baseline Comparison}
%     \label{fig:intervention_sample}
% \end{figure}

% \paragraph{\name Intervention Analysis}

\paragraph{Setting.}
% \hao{we need to clarify why we cannot use EM here}
% In this section, using Exact Match (EM) with gold answers is not feasible because LM outputs often differ in format (e.g., "year before month") or provide partial matches (e.g., just the year). Preliminary results on Llama3-8B showed 0.0 EM accuracy for both base and intervened outputs on the test split of NQ dataset, making it uninformative for evaluating improvements. Instead, we aim to assess whether interventions produce more accurate outputs than the base model, even if they don't perfectly align with the gold answers. 
Our preliminary experiments indicate that Exact Match (EM) fails to capture the nuanced improvements introduced by interventions. EM's binary nature overlooks partial corrections, which are common in our setting. For instance, if the gold answer is \textit{May 2024}, the base model outputs \textit{2025}, and the intervened model outputs \textit{May}, EM considers both as equally incorrect. However, the intervened output is clearly closer to the gold answer. This limitation makes EM uninformative for evaluating interventions that move outputs toward greater factual accuracy, even if they don't perfectly align with the gold answer.
Hence, following recent works \citep{raju2024constructing,chen2024mj}, we employ GPT-4o~\citep{openai2024gpt4technicalreport} as the evaluator to assess for factuality. See Appendix~\ref{appendix:prompt} for the full prompt. 
Human evaluation performed by the authors indicate that there is a substantial agreement between GPT-4o and human judgement, with a Cohen's Kappa of 0.6 (substantial agreement), justifying our choice of using GPT-4o as an automatic evaluation metric.
For the stochastic \intervention, we sample 1, 10, 20, and 30 different \(\boldsymbol{\epsilon}\), and apply the interventions;
we then use \classifier to select the intervened hidden state that leads to the highest probability by \classifier,
which is then used for onward decoding.\footnote{A higher probability by \classifier indicates the hidden state is more likely to lead to a factual output.}
We apply the adjustment only to the first decoding step,
modifying ${\vh}_{I}^{(l)}$ to $\widetilde{\vh}_{I}^{(l)}$ when the classifier's confidence $\alpha$ is less than or equal to $0.3$.

\paragraph{Results.}
Figure~\ref{fig:intervention_results} summarizes the performance of \name's intervention performance on the NQ-open dataset, including both the deterministic and stochastic variants, with greedy decoding.
The intervened LMs consistently outperform the base LMs, with a higher proportion of wins favoring the adjusted outputs. The deterministic intervention consistently achieves a win rate of at least 60\% in all cases, while  without interventions (Base), the LMs show significantly lower performance, with wins as low as 34\%. On average, the winning rate of LMs with intervention across all intervention models is 34.4\% higher than that of the base LMs.
A similar trend is observed on the MedMCQA dataset; results are provided in \ref{appendix:intervention_medmcqa_results}.

The results demonstrate that both deterministic and stochastic intervention models improve the factuality of LM's outputs. These finding suggest that, we can mitigate the hallucination even before it shows up in the generation of the LM.
Additional qualitative examples are presented in Appendix~\ref{appendix:qualitative_samples}.
% These findings suggest that adjusting hidden states preemptively can lead to notable improvements in mitigating hallucinations.

We note that direct comparison with decoding-based methods such as \citep{chuang2024doladecodingcontrastinglayers} may not be directly comparable, as they assess factuality after tokens have been generated, whereas \name predicts hallucination risk before generation begins. 
Similarly, existing mitigation methods typically assess factuality and apply corrections on the generated output at different stages, including the hidden states \citep{zhang2024truthxalleviatinghallucinationsediting, li2024inferencetimeinterventionelicitingtruthful}, while \name operates purely on input-conditioned hidden states.
Due to this fundamental difference, we focus on preemptive detection rather than methods that leverage the full generated output for mitigation. Our goal is to explore the feasibility and potential benefits of early intervention, which may help reduce hallucinations before they occur.

% \hao{some conclusions to draw from these results}

%%%%%%%%%%%%%%%%%%%%%%%%%%%%%%%%%%%
\definecolor{orange1}{HTML}{FFA500}
\definecolor{blue1}{HTML}{4682B4}
\definecolor{green1}{HTML}{9ACD32}
% \begin{figure}[h]
\begin{figure*}  % Force the figure to stay here
    \centering
    \begin{minipage}{0.5\textwidth}
        \centering
        \begin{tabular}{lll}
        \textcolor{orange1}{\rule{5mm}{2mm}} Intervened wins & \textcolor{green1}{\rule{5mm}{2mm}} Ties & \textcolor{blue1}{\rule{5mm}{2mm}} Base wins
        
        \end{tabular}
    \end{minipage}
    \begin{subfigure}[b]{1.0\textwidth}
        \centering
        \includegraphics[width=0.8\textwidth]{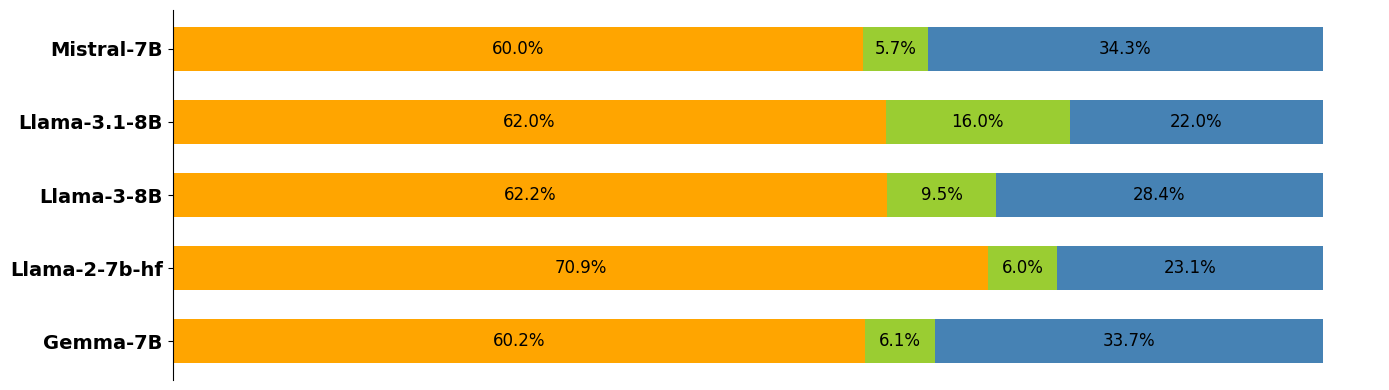}
        \caption{Deterministic \intervention}
        \label{fig:deterministic}
    \end{subfigure}
    \hfill
    \begin{subfigure}[b]{1.0\textwidth}
        \centering
        \includegraphics[width=0.8\textwidth]{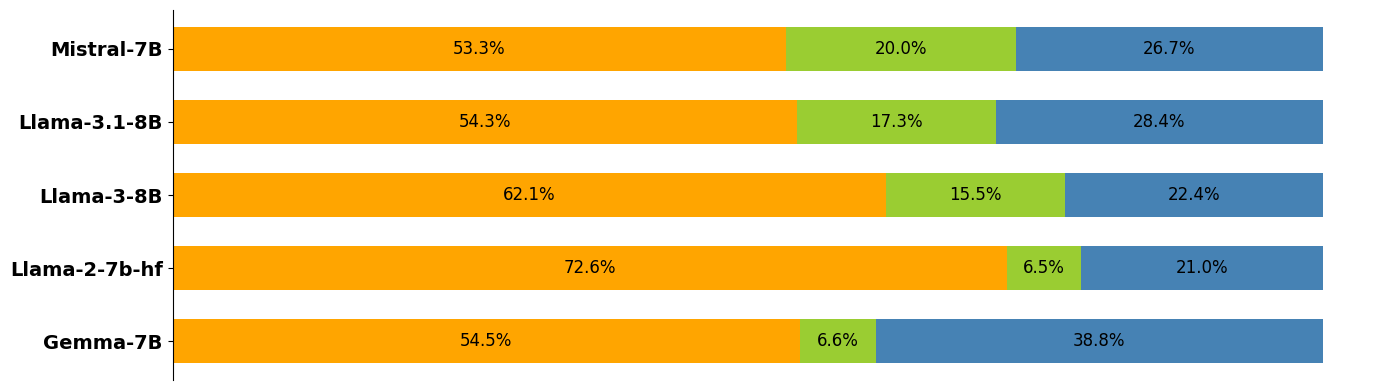}
        \caption{Stochastic \intervention (1 trial)}
        \label{fig:stochastic_1}
    \end{subfigure}
    \hfill
    \begin{subfigure}[b]{1.0\textwidth}
        \centering
        \includegraphics[width=0.8\textwidth]{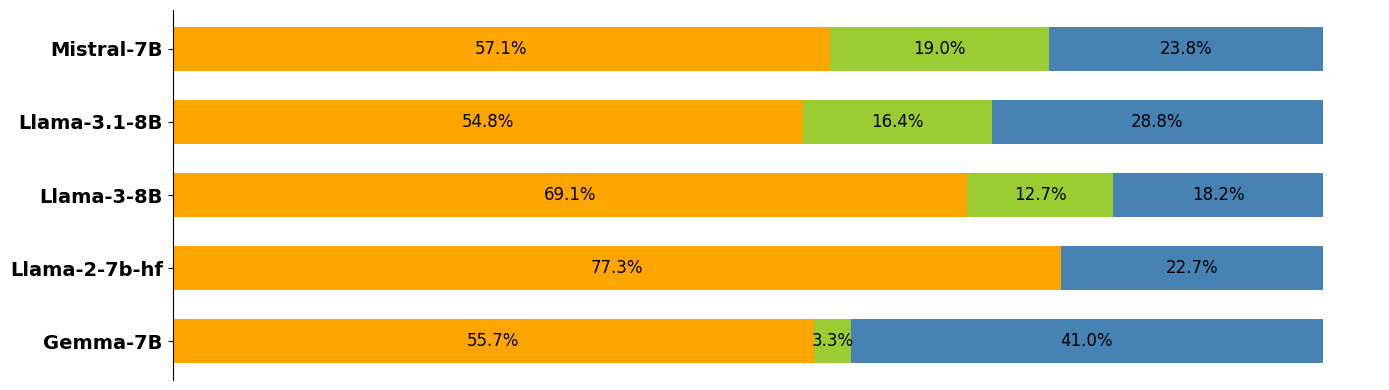}
        \caption{Stochastic \intervention (10 trials)}
        \label{fig:stochastic_10}
    \end{subfigure}
    \hfill
    \begin{subfigure}[b]{1.0\textwidth}
        \centering
        \includegraphics[width=0.8\textwidth]{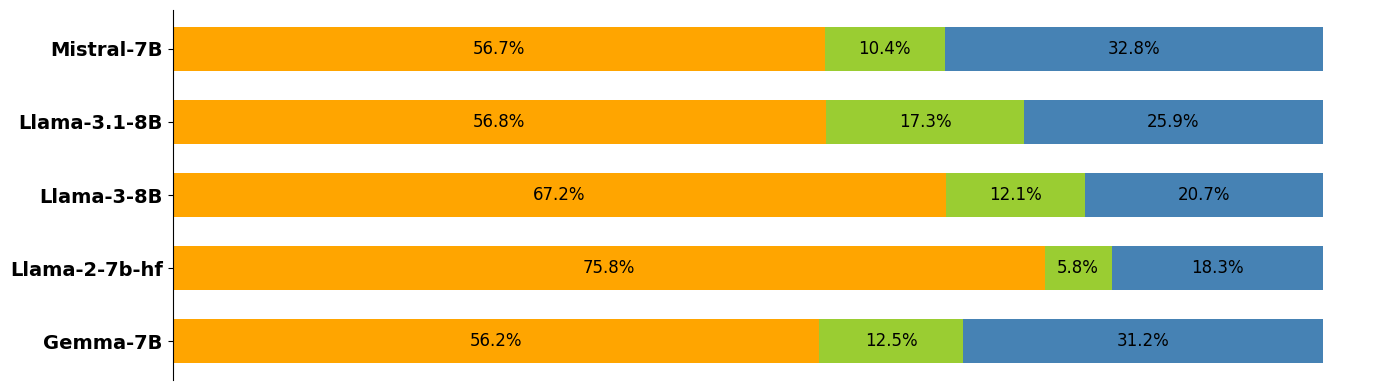}
        \caption{Stochastic \intervention (20 trials)}
        \label{fig:stochastic_20}
    \end{subfigure}
    \hfill
    \begin{subfigure}[b]{1.0\textwidth}
        \centering
        \includegraphics[width=0.8\textwidth]{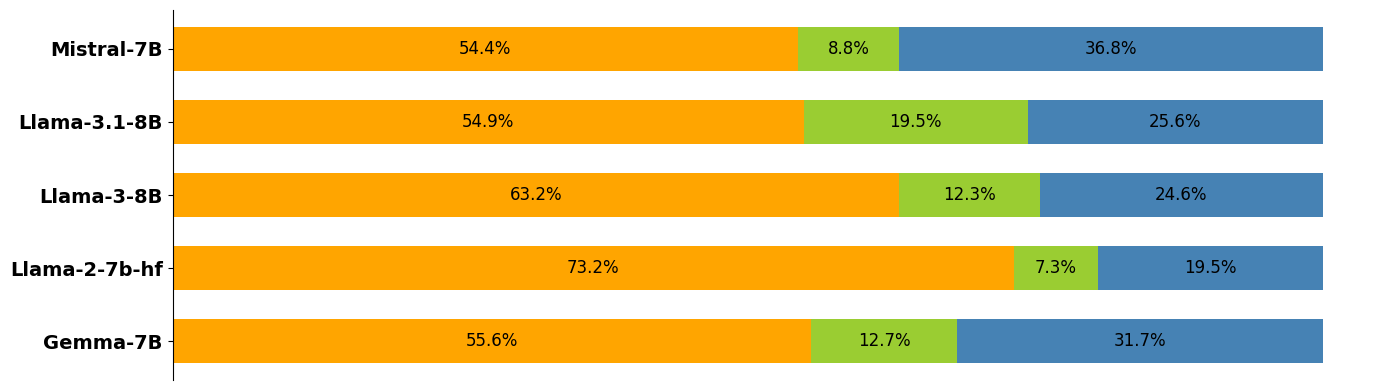}
        \caption{Stochastic \intervention (30 trials)}
        \label{fig:stochastic_30}
    \end{subfigure}
    \caption{Comparison of \name's intervention models. The stochastic model resamples \(\boldsymbol{\epsilon}\) for  1, 10, 20, and 30 times, \classifier used to select the intervened hidden state that leads to the highest probability by \classifier.
    Green color indicates tie, orange for the intervened LM wins, and blue for the base LM wins. (\S\ref{subsec:intervene_results}).
    }
    \label{fig:intervention_results}
    % \vspace{-.5cm}
\end{figure*}

\section{Additional Experiments}\label{sec:additional}
% \deema{these points can be integrated with experiments findings}
% \neeraja{Yes ok, but we can keep the experiment for results across modes here as ablation.}

% \hao{add a quick roadmap} \neeraja{done}
% \hao{i'd reorder things---time overhead first, emebedding second, then different aggregations.} \neeraja{done}

% \hao{remnember to mention the time overhead in the abstract and intro}
% \neeraja{done}
In the following section, we first evaluate the inference time overhead (\S\ref{time_eval_sec}).
Next, we investigate the role of word embedding layers (\S\ref{subsec:word_embedding_exp}). Other experiments are detailed in (\S\ref{app:ablation})
% Experiments that conduct classification experiments to analyze performance across different modes of aggregation are in (\S\ref{subsec:modestypes}). 

\subsection{\name's Time Overhead} 
\label{time_eval_sec}

\begin{table}[!t]
    \centering
\adjustbox{max width=\linewidth}{%
    \begin{tabular}{@{}lc@{}} % Two columns: left-aligned and center-aligned
        \toprule
        \textbf{LM} & \textbf{Inference Time Increase} ($\Delta$) \\ 
        \midrule
        Llama-2-7B   & 0.94\% \\
        Llama-3-8B   & 1.71\% \\
        Llama-3.1-8B & 0.96\%  \\
        \bottomrule
    \end{tabular}}
    \caption{LMs inference time overheads over three runs per LM. 
    The average relative increase in inference time is approximately 1.2\%, showing minimal impact on inference performance
    \S\ref{time_eval_sec}).
    % \hao{suggest changing this to relative}
    % \deema{done}
    % \hao{what about memory overhead?}
    % \deema{working on the memory overhead, guess we dont have time =(}
    }
    \label{tab:lm_intervention}
\end{table}
Both \classifier and \intervention are lightweight and should incur minimal inference overhead.
We confirm this across three models: Llama-2-7B, Llama-3-8B, and Llama-3.1-8B. For each model, the average inference time was measured both with and without \name over three runs, each processing 400 few-shot prompts. 
The results are summarized in Table \ref{tab:lm_intervention}. 
% Figure \neeraja{need to add} %\ref{fig:time_comparison}. 
\name introduces a negligible overhead. We see that the result is consistent over models. This negligible overhead is a promising factor for scaling the experiments or integrating it into the existing LMs' pipelines.
%\hao{i 'm confused why this has anything to do with training}.
% The difference in the timing attributes to the fact, that 

% \begin{table}[htbp]
%     \centering
%     \begin{tabular}{lcc}
%         \toprule
%         \textbf{LM} & \textbf{LM Intervention State} & \textbf{Average Inference Time (min)} \\
%         \midrule
%         llama-2-7B   & Base  &  \\
%         llama-2-7B   & \name  &  18.856\\
%         \cmidrule(lr){1-3}
%         llama3-8B    & Base  &  \\
%         llama3-8B    & \name  &  15.233\\
%         \cmidrule(lr){1-3}
%         llama3.1-8B  & Base  &  \\
%         llama3.1-8B  & \name  &  16.762\\
%         \bottomrule
%     \end{tabular}
%     \caption{Comparison of LMs Inference Time Overheads}
%     \label{tab:lm_intervention}
% \end{table}

\subsection{\texorpdfstring{\classifier} CClassifies the Hidden States Rather than the Questions} 
\begin{table}[!t]
\centering
\adjustbox{max width=\linewidth}{%
\renewcommand{\arraystretch}{1.2} % Adjust row spacing
\setlength{\tabcolsep}{4pt}  
\begin{tabular}{@{}
    l
    >{\centering\arraybackslash\columncolor[HTML]{A1D1F0}}m{3em}
    >{\centering\arraybackslash\columncolor[HTML]{B0BBFA}}m{3em}
    >{\centering\arraybackslash\columncolor[HTML]{FCEFB7}}m{3em}
    >{\centering\arraybackslash\columncolor[HTML]{F4E9E4}}m{3em}
    >{\centering\arraybackslash\columncolor[HTML]{9BDEB8}}m{3em}
}
\toprule
& \multicolumn{4}{c}{\textbf{Preemptive}} \\
\cmidrule(lr){3-6} % Covers columns -1, -2, and -3 only
\textbf{LM} 
& \textbf{I+O} & \textbf{I} & \textbf{-1} & \textbf{-2} & \textbf{-3} \\
\midrule

Llama-2-7B & 
63.9 & 52.3 & 55.3 & 54.9 & 55.6 \\

\bottomrule
\end{tabular}}
\caption{
Results for the word embedding layer of Llama-2-7b on MedMCQA dataset. (\S\ref{subsec:word_embedding_exp}). The table shows classification accuracy of approximately 50\%, indicating no influence of the question difficulty or type on the preemptive hallucination results shown in Table \ref{tab:classification}.}
% \label{tab:classification_prefix}
% \vspace{-.3cm}
\label{tab:wordembeddinglayer}
\end{table}

% Here, we pose the question of whether the intrinsic difficulty of the question contributes to the model's tendency for hallucination classification. As shown in Table \ref{tab:wordembeddinglayer} we calculate the results for the word embedding layer of the model. The accuracies across the table are close to 50 percent. This implies that there is no bias in selecting the questions (prefix) itself while doing our experiments.
\label{subsec:word_embedding_exp}
One possible explanation for \classifier's strong preemptive hallucination detection is that it might be classifying the input questions rather than the LMs' hidden states, since intuitively, more difficult questions could lead to a higher chances of hallucinations by the LMs. However, our results indicate that it is the LMs' hidden states, rather than the questions themselves, that drive the success of \classifier.

% In this section, we explore whether the inherent difficulty of the question prompted to the model influences the hallucination classification scores. 
Table \ref{tab:wordembeddinglayer} summarizes the test accuracies for an \classifier trained and tested on the word embedding layer of Llama-2-7B, before any contextualization by the LM.
Across the board, the accuracies are close to 50\% random guess. 
This confirms that the model is not skewed towards favoring a certain type of question over another while doing the classification. The difficulty of the question is hence, not a contributing factor to the accuracy calculated by classifying the hidden states.

\section{Conclusion}
% In conclusion, we observe that the hidden states of the model preemptively know the misbehaviour present in the model. This finding helps us in detecting the misbehaviour early and steer the model towards a more factual generation. 

% In conclusion, in this work, we see that the hidden states have a lot of information encoded in them that we can leverage. The hidden states are able to predict the misbehaviour of the model preemptively even before the hallucinated text is generated. 
% Using this, we help steer the generation towards a more factual generation. This pipeline can be adopted while training a model from scratch so that it does not generate hallucinated content. We show the feasibility of this study by showing the results on various datasets and multiple varieties of models.
In conclusion, \name demonstrates that the LMs' hidden states encode rich information that can be used to predict hallucination preemptively, even before they appear in the generated output.
% In \name, 
Leveraging this insight, we develop a preemptive detection and intervention mechanism that steers the LM's generation towards more factual outputs, once the hallucination is likely to occur.
We achieve a preemptive hallucination detection accuracy of more than 70\%, and an average of 34.4\% more factual output by LMs supported by \name, compared to the base LMs.
% \name empirically proves the significant potential of utilizing the internal working of LMs, through learning lightweight models for hallucination detection and mitigation, introducing a negligible average overhead of 1.2\% increase in the inference time.
We empirically prove the significant potential of utilizing the internal working of LMs, through learning lightweight models for hallucination detection and mitigation, introducing a negligible average overhead of 1.2\% increase in the inference time.

\section{Limitations and Future work}
% Going ahead, we want to investigate how this pipeline works for out-of-domain data. The current pipeline does not work effectively for out of domain data. We have also looked at datasets which are in distribution to the model being used. We would like to see the effects of this on other data. \neeraja{Is this necessary?}
% We have only looked at hidden states as an internal component in doing the classification to predict the factuality of a sentence. Looking at other internal components of the model would also be a future direction. We also want find out a more robust and consistent way to steer the generation of the model towards the truth.

% limitation is that layer have differnet affects on the noise and also the tao plays a factor - done

While our current experiments focus on QA tasks due to their structured evaluation, the core idea behind \name leveraging hidden states for preemptive hallucination detection and mitigation could extend to other generation tasks. For tasks like summarization or dialogue systems, we anticipate similar patterns in hidden states when the model is about to generate hallucinated content. However, pinpointing where the hallucination starts to analyze preceding hidden states is challenging and presents a more nuanced problem. Therefore, alternative assessment metrics, such as faithfulness scoring or costly human annotation, would be needed instead of exact match evaluation.

Additionally, we have only looked at the hidden states as an internal component for classification to predict the factuality of a sentence. Exploring other LM's internal components presents a potential direction for future work. This pipeline also would not work for Black-box LMs.
% With our detection and intervention experiments, we see that different layers in the LM have varying effects. 
% The classifier \classifier and the intervention model \intervention, are sensitive to the hyperparameters selected. 
% To solve this, we want find out a more robust and consistent approach that is less sensitive to varying hyperparameters, to steer the generation of the model towards the truth.

% Our architecture does not give good results for out of domain data. We want to make it more robust and generalizable going ahead. The datasets that we have used also have the same distribution as that of the model. Extending this to other data and seeing the effects can be a future direction.

% Going ahead, we aim to build a pipeline that is more generalizable and is applicable to a variety of domains.
% Expanding our evaluation to include diverse datasets with different distributions could provide valuable insights and a potential future direction for improving model generalizability.
% \name has demonstrated promising results on question-answering tasks. A natural extension would be to explore its application in more demanding tasks, such as dialogue-based and long-form generation, where accurately identifying the onset of hallucinations is especially challenging.

\section{Ethics Statement}
In developing the \name framework, we committed to advancing the ethical use and dependability of large language models (LLMs). We recognize that as LLMs increasingly permeate various aspects of life, ensuring their reliability and truthfulness in generating content is paramount. Thus, our research focused on preemptively detecting factual inaccuracies, aiming to mitigate potential misinformation spread and reduce the propagation of biases present in training data. Furthermore, we meticulously avoided using any data that could potentially compromise individual privacy or confidentiality and ensured our data handling procedures comply with relevant data protection regulations.

% \input{acknowledgement}

% \bibliography{custom}
\nocite{llama2}
\nocite{mistral}
\nocite{gemma}
\bibliography{iclr2025_conference}

\appendix
\section{Dataset size}
\label{app:datasetsize}
\begin{table*}[htbp]
    \centering
    \begin{tabular}{@{} lrrrr@{}}
        \toprule
        \textbf{Dataset} & \textbf{Total Size} & \textbf{Train (70\%)} & \textbf{Validation (15\%)} & \textbf{Test (15\%)} \\
        \midrule
        NQ-Open~\citep{lee-etal-2019-latent} & 6,666 & 4,666 & 1,000 & 1,000 \\
        MMLU~\citep{mmlu} & 1,844 & 1,292 & 276 & 276 \\
        MedMCQA~\citep{pal2022medmcqa} & 2,000 & 1,400 & 300 & 300 \\
        GSM8K~\citep{cobbe2021trainingverifierssolvemath} & 1,040 & 728 & 156 & 156 \\
        \bottomrule
    \end{tabular}
    \caption{Dataset splits and sizes for training the hallucination classifier \classifier over the LMs' hidden states (\S\ref{subsec:classifier}). }
    \label{tab:data_sizes}
    \vspace{-.3cm}
\end{table*}
\section{Factual Assessment Prompt}
\label{appendix:prompt}

To assess factual accuracy, we use GPT-4o~\citep{openai2024gpt4technicalreport} as the evaluator. To reduce stochasticity in the prompting process, we set the temperature to \(1 \times 10^{-14}\) and top\_p to \(1 \times 10^{-17}\). The prompt used for evaluation is as follows:
\begin{tcolorbox}
\textbf{System:} 
\textit{You are an expert evaluator with an access to Google Search. Your task is to evaluate two responses to a question for factual accuracy. For this task, 'Factual accuracy' refers to the correctness and relevance of the information, aligned with facts accepted or verified as recent as 2021. Ignore stylistic differences, length, opinions, or phrasing unless they change the factual meaning. Supported by your Google Search results, decide which response, if any, is correct. Answer 'first' if the first response is the only correct response, 'second' if the second response is the only correct response, 'both' if both responses are correct, or 'neither' if neither response is correct or if the information provided is ambiguous or insufficient for making a decision, You should favor the response that shows uncertainty if the other response is incorrect. Then, in a new line, briefly explain the reason.}

\textbf{User}: 
\emph{Question:} who played first game in world cup 2018?
\emph{First Response:}  Russia vs Saudi Arabia
\emph{Second Response:} Brazil vs Germany.
\end{tcolorbox}

% need to add the answers?
\section{Experiments for classification}
\subsection{Hidden Representation Classification Analysis}
\label{Classification_analysis}
% \hao{this whole section fits better  in the appendix. we don't need to talk about across layers/ pooling here} \neeraja{done}

\begin{figure}[ht]
    \centering
    % \begin{subfigure}[b]{0.6\textwidth}
    %     \centering
    %     \includegraphics[width=\textwidth]{figures/prefixcompare.png}
    %     \caption{Comparison of prefix experiments over the mode of aggregation}
    %     \label{fig:prefix-over-modes}
    % \end{subfigure}
    % \hfill
    \begin{subfigure}[b]{0.5\textwidth}
        \centering
        \includegraphics[width=\textwidth]{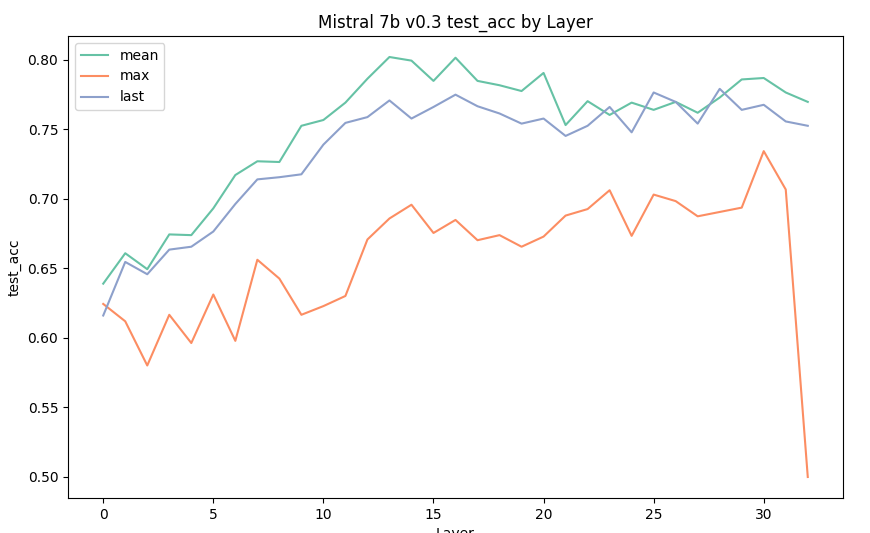}
        \caption{Test Accuracy by Layer for all modes for the Mistral-7b}
        \label{fig:gemma-7b_test_acc_by_layer}
    \end{subfigure}
    \hfill
    \begin{subfigure}[b]{0.5\textwidth}
        \centering
        \includegraphics[width=0.98\textwidth]{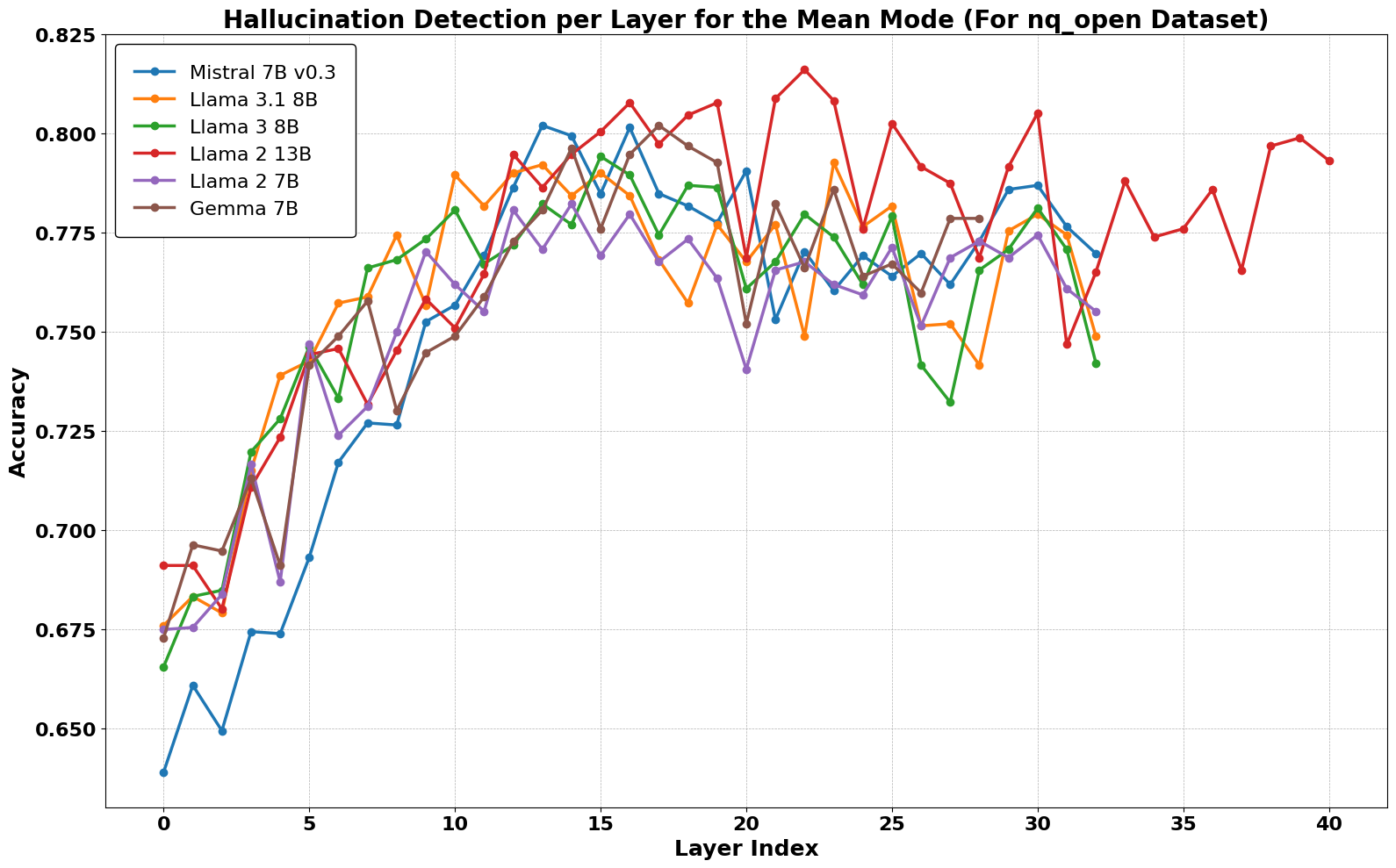}
        \caption{Accuracies for entire sentence across models and layers.}
        \label{fig:example}
    \end{subfigure}
    
    \label{fig:cls_stuff}
\end{figure}

Given the datasets and models described above, for every layer in a model we train a corresponding classifier on hidden states of that respective layer. We use three
modes for aggregating the hidden states before passing them to the classifier: mean pooling, max
pooling and taking the last token in the hidden states. Figure~\ref{fig:example} illustrates the accuracy of hallucination detection of the classifiers for the entire sequence, using the mean token representation for aggregation. As shown, the accuracy across all evaluated models mostly exceeds 0.75, indicating a robust capability to identify hallucinations. This high level of performance underscores the efficacy of the hidden state representations in distinguishing factual accuracies within generated content. As seen in the figure, we see that the accuracy peaks for the middle layers. The best performing layer per model and dataset is shown in Table \ref{tab:best_layers}.
%Our experiments also explore taking the element-wise max over hidden states, or taking the last one as the input to \classifier,
%and we find that they slightly underperformed taking the mean.
% as explained in section \ref{Classification_analysis}

% \hao{move to experiments: Our experiments also explore taking the elementwise max over hidden states, or taking the last one as the input to \classifier,
% % and find they slightly underperform taking the average as explained in section \ref{Classification_analysis}.} \neeraja{done} 

%\par

 Therefore, for all models we calculate the test accuracy across all layers and all modes of aggregation. Quantitative results are shown in the first column of Table \ref{tab:classification}. Given the three modes of aggregation, we see that mean pooling gives the best results in most cases. Figure \ref{fig:gemma-7b_test_acc_by_layer} shows the test accuracy per layer and mode.  

 \textbf{ Classifier Setup}: The classifier \classifier is two-layer MLP with ReLU activations and BCELoss, trained using an Adam optimizer with a learning rate of \(10^{-4}\) with a dropout rate of 0.1. We train all classifiers for 50 epochs and apply early stopping based on the validation accuracy.

 \textbf{Intervention Setup}: The intervention model \intervention is a three-layer RELU MLP. It is trained using an Adam optimizer. MSE loss is used between the altered hidden state and original hidden state. We train it for 100 epochs.

%%%%%%
%and the intervention model is a three-layer MLP with MSELoss. We apply early stopping to the classifier based on the validation accuracy.
%%%%%%%

% \input{tables/classification}
% \begin{figure}[h]
%     \centering
%     \includegraphics[width=0.5\textwidth]{figures/cls_plots/nq_open.png}
%     \caption{Accuracies for entire sentence across models and layers.
%     % \hao{these accuracies by layer figures fit better in the appendices}
%     }
%     \label{fig:example}
% \end{figure}

% \begin{figure}[ht]
%     \centering
%     \includegraphics[width=0.5\textwidth]
%     {figures/cls_plots/gemma_7b_test_acc_by_layer.png}
%     \caption{Test Accuracy by Layer for Gemma-7b}
%     \label{fig:gemma-7b_test_acc_by_layer}
% \end{figure}

%   \begin{figure}
%         \centering
%         \includegraphics[width=0.4\linewidth]{figures/prefixcompare.png}
%         \caption{Comparison of prefix experiments over the mode of aggregation}
%         \label{fig:prefix-over-modes}
%     \end{figure}

% \documentclass{article}
% \usepackage{booktabs}

% \begin{document}

\begin{table}[ht!]
\centering
\small
\begin{tabular}{lccc}
\toprule
\textbf{LM} & \textbf{NQ} & \textbf{MMLU} & \textbf{MedMCQA} \\
\midrule
Llama-2-7b-hf       &  14    &  16    &   14        \\
Llama-2-13b-hf      &  22    &  15   &    14     \\
Llama-3-8B          &  15    &  17    &   11      \\
Llama3.1-8B         &  23    &  14    &   15      \\
Mistral-7B          &  13    &  -    &    12     \\
Gemma-7B            &  17    &  17    &   18      \\
Llama2-7B-chat  &  14    &  13    &   14      \\                                     
Llama2-13B-chat      &  19    &  14    &   16      \\                               
Llama3-8B-Instruct   &  15    &  15    &   12     \\                                
Llama-3.1-8B-Instruct  &  15    &  13    &   14      \\
Llama3-70B-Instruct  &  74    &  35    &   75      \\
Mistral-7B-Instruct  &  18   &  15    &   18      \\
Gemma-7B-it  &  16    &  18    &   18      \\
% Llama3-70B & 76 & - & -\\

\bottomrule
\end{tabular}
\caption{Best Performing layer per model and dataset}
\label{tab:best_layers}
\end{table}

% \end{document}

% \begin{figure}
%     \centering
%     \includegraphics[width=0.8\linewidth, angle=-90]{figures/trial_10_updated.pdf}
%     \caption{Enter Caption}
%     \label{fig:enter-label}
% \end{figure}

%%%%%%%%%%%%%%%%%%%%%%%%%%%%%%%%%%%%%%%%%%%%%%%%%%%%%%%%%%%%%

% \begin{table}[ht]
% \centering
% \caption{Performance Metrics for Llama 3.1-8B Model After Intervention}
% \label{tab:metrics}
% \begin{tabular}{lc}
% \toprule
% \textbf{Metric}     & \textbf{Value} \\
% \midrule
% Precision           & 0.987          \\
% Recall              & 0.571          \\
% F1-Score            & 0.723          \\
% Accuracy            & 0.623          \\
% \bottomrule
% \end{tabular}
% \end{table}

%%%%%%%%%%%%%%%%%%%%%%%%%%%%%%%%%%%%%%%%%%%%%%%%%%%%%%%%%%%%
\subsection{Additional Experiments}
\begin{table}[!t]
\centering
\adjustbox{max width=.48\textwidth}{%
\renewcommand{\arraystretch}{1.2} % Adjust row spacing
\setlength{\tabcolsep}{4pt}  
\begin{tabular}{@{}
    l
    >{\centering\arraybackslash\columncolor[HTML]{A1D1F0}}m{3em}
    >{\centering\arraybackslash\columncolor[HTML]{B0BBFA}}m{3em}
    >{\centering\arraybackslash\columncolor[HTML]{FCEFB7}}m{3em}
    >{\centering\arraybackslash\columncolor[HTML]{F4E9E4}}m{3em}
    >{\centering\arraybackslash\columncolor[HTML]{9BDEB8}}m{3em}
}
\toprule
& \multicolumn{4}{c}{\textbf{Preemptive}} \\
\cmidrule(lr){3-6} % Covers columns -1, -2, and -3 only
\textbf{LM} 
& \textbf{I+O} & \textbf{I} & \textbf{-1} & \textbf{-2} & \textbf{-3} \\
\midrule

Llama3-70B & 
78.0 & 77.2 & 76.8 & 76.8 & 76.5 \\
Qwen2.5-32B & 
79.8 & 76.3 & 75.9 & 75.5 & 75.9 \\
Llama3.1-70B-Instruct & 
85.3 & 81.0 & 77.6 & 70.9 & 77.2 \\

\bottomrule
\end{tabular}}
\caption{
Results for the gsm8k dataset on the following models}
% \label{tab:classification_prefix}
\vspace{-.3cm}
\label{tab:gsm8k}
\end{table}

\begin{table}[!t]
\centering
\adjustbox{max width=.48\textwidth}{%
\renewcommand{\arraystretch}{1.2} % Adjust row spacing
\setlength{\tabcolsep}{4pt}  
\begin{tabular}{@{}
    l
    >{\centering\arraybackslash\columncolor[HTML]{A1D1F0}}m{3em}
    >{\centering\arraybackslash\columncolor[HTML]{B0BBFA}}m{3em}
    >{\centering\arraybackslash\columncolor[HTML]{FCEFB7}}m{3em}
    >{\centering\arraybackslash\columncolor[HTML]{F4E9E4}}m{3em}
    >{\centering\arraybackslash\columncolor[HTML]{9BDEB8}}m{3em}
}
\toprule
& \multicolumn{4}{c}{\textbf{Preemptive}} \\
\cmidrule(lr){3-6} % Covers columns -1, -2, and -3 only
\textbf{LM} 
& \textbf{I+O} & \textbf{I} & \textbf{-1} & \textbf{-2} & \textbf{-3} \\
\midrule

Qwen2.5-32B & 
92.7 & 90.8 & 86.9 & 87.3 & 74.4 \\
Qwen2.5-32B-Instruct & 
92.7 & 92.1 & 83.7 & 86.2 & 83.8 \\

\bottomrule
\end{tabular}}
\caption{
Results for the mmlu dataset on the following models}
% \label{tab:classification_prefix}
\vspace{-.3cm}
\label{tab:mmlu}
\end{table}

\section{Impact on Nominal Questions}

A key consideration is whether the proposed intervention method impacts nominal, non-hallucinatory questions, particularly given the classifier's false positive rate (FPR). To address this, we conducted a detailed analysis of non-hallucinatory responses generated by the Llama 3.1-8B model on the MedMCQA dataset after intervention.

\begin{table}[ht]
\centering
\caption{Confusion Matrix for Llama 3.1-8B Model After Intervention}
\label{tab:confusion_matrix}
\begin{tabular}{lcc}
\toprule
\textbf{} & \multicolumn{2}{c}{\textbf{Actual}} \\
\cmidrule(lr){2-3}
\textbf{} & \textbf{Positive} & \textbf{Negative} \\
\midrule
\textbf{Predicted Positive} & 672 (TP) & 9 (FP) \\
\textbf{Predicted Negative} & 505 (FN) & 176 (TN) \\
\bottomrule
\end{tabular}
\end{table}

The classifier achieved robust performance metrics: Precision: 0.987, Recall: 0.571, F1-Score: 0.723, and Accuracy: 0.623. Among the 1,362 responses analyzed, 9 cases were identified as false positives, where the intervention occurred despite the absence of hallucination. Importantly, only 2 out of these 9 false positives led to a degradation in the factual quality of the generated output. This finding highlights the minimal disruptive impact of our intervention on non-hallucinatory responses.

Furthermore, interventions correctly enhanced factual outputs in 672 true positive cases, demonstrating the classifier's ability to effectively improve factuality while minimizing unnecessary disruptions. This cautious approach ensures that the vast majority of factual responses remain unaffected, while significant improvements are achieved for hallucinatory responses.

These results alleviate concerns about the classifier's FPR by showing that the proposed method maintains high reliability, minimally impacting nominal questions while effectively enhancing the factuality of hallucinatory outputs.

%%%%%%%%%%%%%%%%%%%%%%%%%%%%%%%

\section{MedMCQA Dataset Intervention Results}
\label{appendix:intervention_medmcqa_results}

\begin{figure*}[ht]
    \centering
    % Legend
    \begin{minipage}{0.5\textwidth}
        \centering
        \begin{tabular}{lll}
            \textcolor{orange1}{\rule{5mm}{2mm}} Intervened wins & 
            \textcolor{green1}{\rule{5mm}{2mm}} Ties & 
            \textcolor{blue1}{\rule{5mm}{2mm}} Base wins
        \end{tabular}
    \end{minipage}
    % Actual Figure
    \vspace{0.5em} % Add slight spacing
    \includegraphics[width=0.8\textwidth]{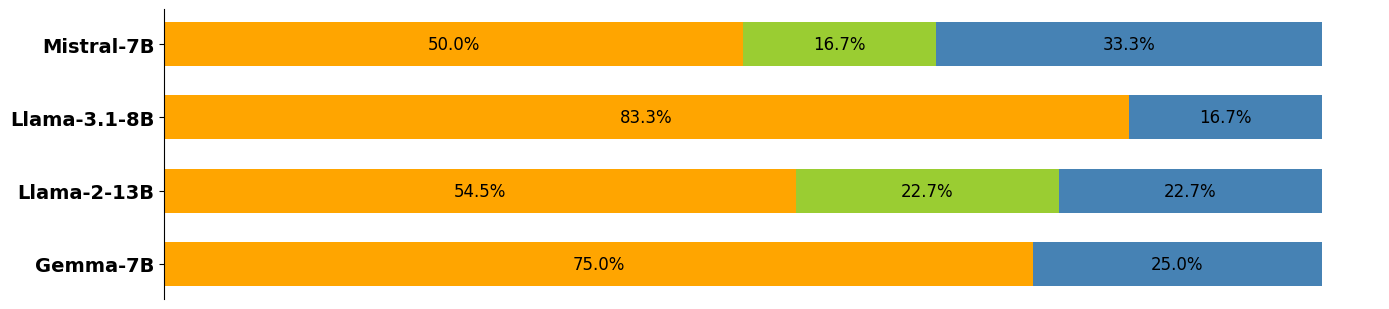}
    \caption{Comparison of \name's deterministic \intervention intervention model on the MedMCQA dataset, under greedy decoding. Green color indicated tie, orange for the intervened LM, and blue for the base LM. Similar to the NQ results, the intervened LMs achieve significantly higher win rates compared to the base LMs. (\S\ref{appendix:intervention_medmcqa_results})}
    \label{fig:medmcqa_results}
\end{figure*}

\paragraph{Results.}
Figure~\ref{fig:medmcqa_results} presents the results of the deterministic \intervention on the MedMCQA dataset. The trends are consistent with those observed for NQ-open: the intervened LMs consistently outperform the base LMs. These results further highlight the effectiveness of \name's intervention model in mitigating hallucinations and improving factuality across several domains.

\section{Qualitative Examples}
\label{appendix:qualitative_samples}

\begin{table*}
\centering
\begin{tabular}{|p{4.5cm}|p{2.5cm}|p{3.5cm}|p{5cm}|}
\hline
\textbf{Question} & \textbf{Base Answer} & \textbf{Answer after Intervention} & \textbf{Explanation} \\
\hline
When was the last time the death penalty was used in PA? & June & (1999) & The last execution in Pennsylvania was in 1999. The second response is correct. \\
\hline
How many Black Eyed Pea restaurants are there? & Over one hundred & 0 & The Black Eyed Pea restaurant chain has significantly reduced its number of locations, and as of recent years, it has closed all of its locations. Therefore, the second response is correct. \\
\hline
When does Jess come back on New Girl? & Season six & Season five & Jess returns to New Girl in season five after a brief absence. \\
\hline
Who plays Angela's father on Boy Meets World? & William Daniels & (Unknown) & The character of Angela's father on Boy Meets World is played by Julius Carry, not William Daniels. William Daniels played Mr. Feeny on the show. The second response is correct in indicating uncertainty as it does not provide incorrect information. \\
\hline
\end{tabular}
\caption{Qualitative Examples. Comparison of base and intervened answers with their explanations from GPT4o justifying its choice.}
\label{tab:comparison_table}
\end{table*}

\section{Generalization results}
\label{Generalization appendix}

\begin{table*}[ht]
    \centering
     % Reduce text size
    \renewcommand{\arraystretch}{0.9}  % Reduce row spacing
    \setlength{\tabcolsep}{4pt}  % Reduce column spacing
    \begin{tabular}{{@{}
    lll
    >{\centering\arraybackslash\columncolor[HTML]{A1D1F0}}m{3em}
    >{\centering\arraybackslash\columncolor[HTML]{B0BBFA}}m{3em}
    >{\centering\arraybackslash\columncolor[HTML]{FCEFB7}}m{3em}
    >{\centering\arraybackslash\columncolor[HTML]{F4E9E4}}m{3em}
    >{\centering\arraybackslash\columncolor[HTML]{9BDEB8}}m{3em}
}}
        \toprule
        \textbf{Model used} & \textbf{Training Data} & \textbf{Test Data} & \textbf{I + O} & \textbf{I} & \textbf{I - 1} & \textbf{I - 2} & \textbf{I - 3} \\
        \midrule
        \multirow{3}{*}{Llama2-7b-hf} 
        & \multirow{3}{*}{MMLU + NQ-Open} & NQ-Open  & 75.6 & 70.8 & 67.0 & 67.4 & 68.6 \\
        &  & MMLU      & 80.0 & 72.3 & 70.7 & 68.7 & 68.9 \\
        &  & MedMCQA   & 63.2 & 58.8 & 58.7 & 59.2 & 58.5 \\
        \bottomrule
    \end{tabular}
    \caption{\textcolor{purple}{Test accuracy results of Llama2-7b-hf on different test datasets and out-of-domain data.}}
    \label{tab:gen-ood}
\end{table*}

\begin{table*}[ht]
    \centering
    \small
    \begin{tabular}{{@{}
    ll
    >{\centering\arraybackslash\columncolor[HTML]{A1D1F0}}m{3em}
    >{\centering\arraybackslash\columncolor[HTML]{B0BBFA}}m{3em}
    >{\centering\arraybackslash\columncolor[HTML]{FCEFB7}}m{3em}
    >{\centering\arraybackslash\columncolor[HTML]{F4E9E4}}m{3em}
    >{\centering\arraybackslash\columncolor[HTML]{9BDEB8}}m{3em}
}}
        \toprule
        \textbf{Trained on} & \textbf{Tested on} & \textbf{I + O} & \textbf{I} & \textbf{I - 1} & \textbf{I - 2} & \textbf{I - 3} \\
        \midrule
        \multirow{2}{*}{Llama 2 7B} & Llama 3.1 8B  & 48.4 & 48.4 & 48.4 & 48.4 & 48.4 \\
                                    & Mistral V0.3 7B & 51.5 & 51.5 & 51.5 & 51.5 & 51.5 \\
        \midrule
        \multirow{2}{*}{Llama 3.1 8B} & Llama 2 7B  & 47.7 & 48.3 & 48.3 & 48.4 & 48.3 \\
                                      & Mistral V0.3 7B & 51.5 & 51.5 & 51.5 & 51.5 & 51.5 \\
        \midrule
        \multirow{2}{*}{Mistral V0.3 7B} & Llama 2 7B  & 48.8 & 48.4 & 48.4 & 48.4 & 48.4 \\
                                         & Llama 3.1 8B & 48.5 & 48.4 & 48.4 & 48.4 & 48.4 \\
        \bottomrule
    \end{tabular}
    \caption{\textcolor{purple}{Performance results of different models on various test datasets.}}
    \label{tab:multi-model:ood}
\end{table*}

\section{Ablation Study}
\label{app:ablation}

\begin{table*}[!t]
\centering
\begin{adjustbox}{max width=\textwidth}
\renewcommand{\arraystretch}{1.2} % Adjust row spacing
\setlength{\tabcolsep}{4pt}       % Adjust column padding
\begin{tabular}{@{}
    l
    >{\centering\arraybackslash\columncolor[HTML]{A1D1F0}}m{2.5em} % I+O Blue
    >{\centering\arraybackslash\columncolor[HTML]{B0BBFA}}m{2.5em} % I Light Blue
    >{\centering\arraybackslash\columncolor[HTML]{FCEFB7}}m{2.5em} % -1 Yellow
    >{\centering\arraybackslash\columncolor[HTML]{F4E9E4}}m{2.5em} % -2 Beige
    >{\centering\arraybackslash\columncolor[HTML]{9BDEB8}}m{2.5em} % -3 Green
    m{0.01em}
    >{\centering\arraybackslash\columncolor[HTML]{A1D1F0}}m{2.5em} % I+O Blue
    >{\centering\arraybackslash\columncolor[HTML]{B0BBFA}}m{2.5em} % I Light Blue
    >{\centering\arraybackslash\columncolor[HTML]{FCEFB7}}m{2.5em} % -1 Yellow
    >{\centering\arraybackslash\columncolor[HTML]{F4E9E4}}m{2.5em} % -2 Beige
    >{\centering\arraybackslash\columncolor[HTML]{9BDEB8}}m{2.5em} % -3 Green
    m{0.01em}
    >{\centering\arraybackslash\columncolor[HTML]{A1D1F0}}m{2.5em} % I+O Blue
    >{\centering\arraybackslash\columncolor[HTML]{B0BBFA}}m{2.5em} % I Light Blue
    >{\centering\arraybackslash\columncolor[HTML]{FCEFB7}}m{2.5em} % -1 Yellow
    >{\centering\arraybackslash\columncolor[HTML]{F4E9E4}}m{2.5em} % -2 Beige
    >{\centering\arraybackslash\columncolor[HTML]{9BDEB8}}m{2.5em} % -3 Green
}
\toprule
& \multicolumn{5}{c}{\textbf{Mean Pooling}} && \multicolumn{5}{c}{\textbf{Last Token}} && \multicolumn{5}{c}{\textbf{Max Pooling}} \\
\cmidrule(lr){2-6} \cmidrule(lr){8-12} \cmidrule(lr){14-18}
& \cellcolor{white} & \multicolumn{4}{c}{\textbf{Preemptive}} 
&& \cellcolor{white} & \multicolumn{4}{c}{\textbf{Preemptive}} 
&& \cellcolor{white} & \multicolumn{4}{c}{\textbf{Preemptive}} \\
\cmidrule(lr){3-6} \cmidrule(lr){9-12} \cmidrule(lr){15-18}
\textbf{LM} 
& \textbf{I+O} & \textbf{I} & \textbf{-1} & \textbf{-2} & \textbf{-3}
&& \textbf{I+O} & \textbf{I} & \textbf{-1} & \textbf{-2} & \textbf{-3}
&& \textbf{I+O} & \textbf{I} & \textbf{-1} & \textbf{-2} & \textbf{-3} \\
\midrule

Llama3-8B 
& 79.4  & \textbf{75.9} & 73.4 & 72.2 & 71.6 
&& 81.7  & 71.0 & 63.3 & 51.8 & 51.3 
&& 73.1 & 70.5 & 69.9 & 68.8 & 68.9 \\

\bottomrule
\end{tabular}
\end{adjustbox}

\caption{
Comparison of hallucination classification across different aggregation modes for the same layer and LM. We show the results for the Llama3-8B on layer 15. We see that the difference between \textbf{I+O} and \textbf{I} is the least when the mean is the mode of aggregation.}
\label{tab:classificationacrossmodes}
\vspace{-.2cm}
\end{table*}

\subsection{Preemptive Hallucination Detection across various Aggregation Methods 
% \hao{we need to be consistent about capitalization in section/paragraph titles}
}
\label{subsec:modestypes}
We explore three modes for aggregating the hidden states
 % before passing them to the classifier
: mean pooling, max pooling, and taking the last token. We see that the mean pooling shows the best accuracy as shown in Fig \ref{fig:gemma-7b_test_acc_by_layer}. To test how different modes of aggregation work for the preemptive experiments, we compare all the three modes. This is done across the same layer for a the same model. As shown in Table \ref{tab:classificationacrossmodes}, we see that the accuracy of the entire sentence (\colorbox[HTML]{A1D1F0}{\bf I+O}) is similar for last token and mean pooling. However, the drop in the subsequent accuracies is the maximum when last token is used. The maximum accuracy for \colorbox[HTML]{B0BBFA}{\bf I} is when mean pooling is used. Therefore, we use mean pooling as our mode of aggregation in all our experiments. 
\subsection{Ablation Study: Classifier-Based Sampling without Intervention}
\label{subsec:ablation}
\begin{figure*}[ht]
    \centering
    \begin{minipage}{0.5\textwidth}
        \centering
        \begin{tabular}{lll}
        \textcolor{orange1}{\rule{5mm}{2mm}} Sample-\name-CLS wins & \textcolor{green1}{\rule{5mm}{2mm}} Ties & \textcolor{blue1}{\rule{5mm}{2mm}} Base wins
        
        \end{tabular}
    \end{minipage}
    % \begin{subfigure}[b]{1.0\textwidth}
    %     \centering
    %     \includegraphics[width=\textwidth]{figures/Sample-factcheckmate.png}
    %     \caption{Sample-\name \neeraja{pending}}
    %     \label{fig:Sample-factcheckmate}
    % \end{subfigure}
    % \hfill
    % \begin{subfigure}[b]{1.0\textwidth}
    %     \centering
    %     \includegraphics[width=\textwidth]{figures/pca_updated.png}
    %     \caption{PCA}
    %     \label{fig:pca}
    % \end{subfigure}
    % \hfill
    \begin{subfigure}[b]{1.0\textwidth}
        \centering
        \includegraphics[width=0.91\textwidth]{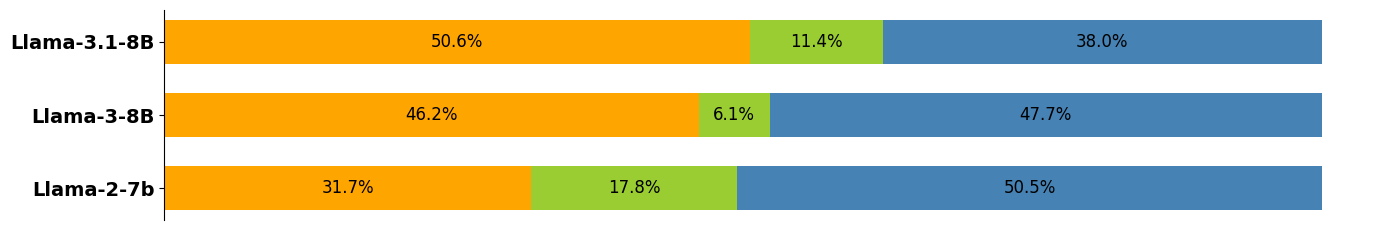}
        \caption{Sample-\name-CLS}
        \label{fig:sampling}
    \end{subfigure}
    \caption{The figure shows the winning rate of the Sample-\name-CLS LM (Orange), the base LM (Blue), and the ties (Green) (\S\ref{subsec:ablation}).}
    \label{fig:intervention_results_comparison}
    \vspace{-.4cm}
\end{figure*}

Sample-\name-CLS refers to a sampling-based decoding approach that leverages the hallucination classifier \classifier component of \name to select the sample with the highest classifier accuracy, without applying intervention. To evaluate its effectiveness, we compare Sample-\name-CLS against the native sampling of the non-intervened LM, using the same random seed for fair comparison, with top-100. % of 100. 
As shown in Figure~\ref{fig:intervention_results_comparison}, while Sample-\name-CLS improves performance in certain cases, it underperforms in others. This highlights the necessity of intervention to effectively mitigate hallucinations and achieve consistent factual improvements.

% \appendix

% \input{sections/appendix}

\end{document}